\documentclass{bmvc2k}
\usepackage{todonotes}

\title{Event-based Non-Rigid Reconstruction \\ from Contours}

\addauthor{Yuxuan Xue}{yuxuan.xue@tuebingen.mpg.de}{1,2}
\addauthor{Haolong Li}{haolong.li@tuebingen.mpg.de}{1}
\addauthor{Stefan Leutenegger}{stefan.leutenegger@tum.de}{2}
\addauthor{Joerg Stueckler}{joerg.stueckler@tuebingen.mpg.de}{1}
\addinstitution{
 Max Planck Institute\\
 for Intelligent Systems, \\
 Tuebingen, Germany
}
\addinstitution{
 Technical University \\
 of Munich,\\
 Munich, Germany
}

\runninghead{Xue, Li, Leutenegger, Stueckler}{Event-based Non-Rigid Reconstruction}

\begin{document}

\maketitle
\begin{abstract}
Visual reconstruction of fast non-rigid object deformations over time is a challenge for conventional frame-based cameras.
   In this paper, we propose a novel approach for reconstructing such deformations using measurements from event-based cameras. Under the assumption of a static background, where all events are generated by the motion, our approach estimates the deformation of objects from events generated at the object contour in a probabilistic optimization framework.
   It associates events to mesh faces on the contour and maximizes the alignment of the line of sight through the event pixel with the associated face.
   In experiments on synthetic and real data, we demonstrate the advantages of our method over state-of-the-art optimization and learning-based approaches for reconstructing the motion of human hands. 
   A video of the experiments is available at \href{ https://youtu.be/upcCVTomFXY}{ https://youtu.be/upcCVTomFXY}.
   \let\thefootnote\relax\footnotetext{Preprint version of the paper accepted for publication in the British Machine Vision Conference (BMVC), 2022}
\end{abstract}
\section{Introduction}
\label{sec:intro}
Event cameras offer a considerable number of advantages in computer vision tasks over conventional cameras, such as low latency, high dynamic range and virtually no motion blur. 
Unlike conventional frame-based cameras that capture images at a fixed rate, event cameras asynchronously measure per-pixel brightness change, and output a stream of events that encode the spatio-temporal coordinates of the brightness change and its polarity. 
While several approaches for event-based cameras have been proposed for optical flow estimation or simultaneous localization and mapping~\cite{gallego_event_survey2018}, only little work has been devoted to non-rigid reconstruction~\cite{nehvi2021_diffevsim, rudnev2021_eventhands}.

In this paper, we present a novel non-rigid reconstruction approach for event-based cameras. 
Our algorithm takes event streams as input and outputs the reconstructed object pose parameters, assuming a low-dimensional parameterized shape template of a deforming object (i.e. hand and body model).
We propose a novel optimization-based method based on expectation maximization (EM). 
Our method models event measurements at contours in a probabilistic way to estimate the association likelihood of events to mesh faces and maximize the measurement likelihood.
The approach is evaluated on synthetic and real data sequences, and improvements over the state-of-the-art optimization and learning-based methods for hand reconstruction are demonstrated.
For generating the sequences, we develop a novel event-based camera simulator for non-rigid deforming objects.
Details of our proposed simulator and a comparison with the state-of-the-art event-based camera simulators~\cite{rebeg2018_esim, nehvi2021_diffevsim, rudnev2021_eventhands} can be found in the supplementary material.

\begin{figure}[tb]
\centering
\centering
\includegraphics[width=0.99\textwidth]{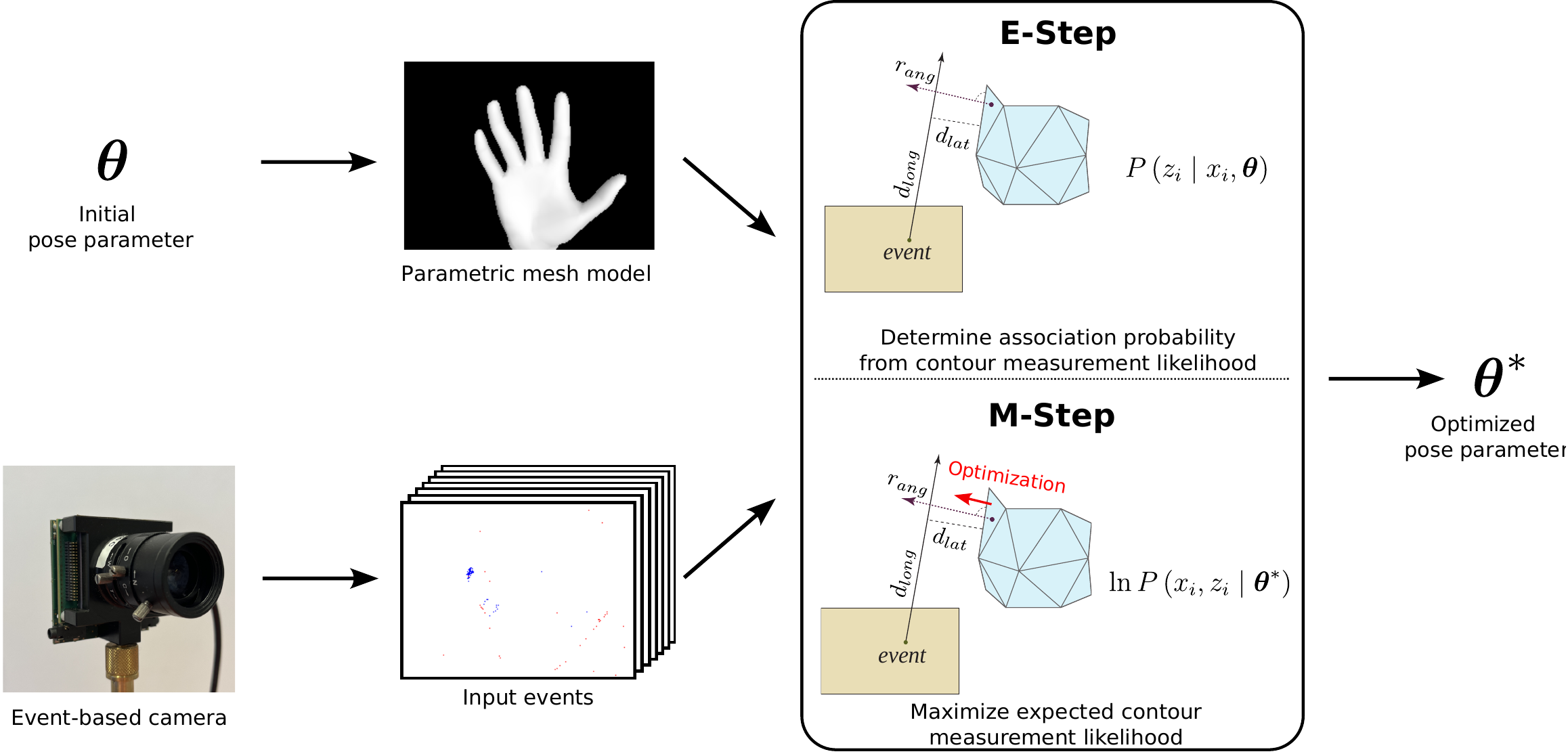}
\caption{Our approach reconstructs non-rigid deformation states of objects from event streams of event-based cameras within an expectation maximization (EM) framework. In the E-step, the association probability of events to contour mesh faces is estimated from the contour measurement likelihood. In the M-step, the expected value of the measurement likelihood over the the association probability  is maximized  for pose parameter $\boldsymbol{\theta^*}$.}
\label{fig:teaser}
\end{figure}
\section{Related Work}
A multitude of approaches has been proposed in recent years for scene reconstruction and rigid tracking using event-based cameras~\cite{kim2016_ev3Drecon,rebecq2017_evvio,vidal2018_ultimateslam}.
Only recently, approaches have been presented that track 3D object motion~\cite{li2021_evobjtrack}.

One can distinguish state-of-the-art approaches into approaches that accumulate disparity space images based on multi-view stereo~\cite{rebecq2018_emvs}, maximize contrast of reprojected events in a reference frame~\cite{gallego2019_focus,stoffregen2019_contrastmaximization}, or apply a generative event measurement model derived from the brightness constancy assumption~\cite{kim2016_ev3Drecon}.

Reconstruction and tracking of non-rigid shapes is a challenging problem in computer vision.
For monocular frame-based cameras, several approaches have been proposed. 
They can be classified into methods that align shape templates (e.g.~\cite{salzmann2009_sharplyfolding,ngo2016_laplacianshape,yu2015_dddtemplate}) or approaches that use regularizing assumptions such as low-rank approximations to achieve non-rigid structure from motion (e.g.~\cite{bregler2000_nrsfm,dai2014_nrsfm,garg2013_varnrsfm,sidhu2020_neuralnrsfm,lamarca2021_defslam}).
RGB-D cameras simplify the task due to the availability of dense depth for which several methods have been proposed recently (e.g.~\cite{newcombe2015_dynamicfusion,bozic2020_neuralnrsfm}).

Non-rigid reconstruction and tracking with event-based cameras has only recently attained attention in the computer vision community.
Nehvi et al.~\cite{nehvi2021_diffevsim} propose a differentiable event stream simulator by subtracting renderings of parametrized hand models. The paper demonstrates the use of the simulator for non-rigid motion tracking from event streams by optimization. 
Rudnev et al.~\cite{rudnev2021_eventhands} train a deep neural network on synthetic event streams to estimate the deformation of a MANO~\cite{mano_hand_model} hand model.
To input the event data into the neural network, they propose to represent the data in local time windows.
Different to these methods, we propose geometric contour alignment in a probabilistic optimization framework.

Some research in recent years has been devoted to event stream simulation~\cite{nehvi2021_diffevsim, rebeg2018_esim, rudnev2021_eventhands}. We propose an event data simulator which generates synthetic events and other data modalities of human body motion, especially of hand deformation by simulating events with an adaptive sampling rate.

\section{Event-based Non-Rigid Reconstruction from Contours}
Our event-based method tracks parameterized non-rigid objects assuming static background. 
Typically, for deforming objects with low texture such as hands or human bodies, the majority of events is generated at the contour between the object and the background.
Hence, we formulate the reconstruction problem in a probabilistic way using a contour measurement model for the events. 
Assuming a known initial state, we optimize for the pose parameters of the parametric object model incrementally from the event stream. 

\subsection{Expectation Maximization Framework}
We formulate the 4D reconstruction problem as maximum-a-posteriori estimation of the model parameters~$\boldsymbol{\theta}$ given the event observations~$\mathbf{x}$ from the event camera
\begin{equation}
    \boldsymbol{\theta}^* = \argmax_{\boldsymbol{\theta}} \ln p(\mathbf{x} \mid \boldsymbol{\theta}) + \ln p(\boldsymbol{\theta}),
\end{equation}
where~$p(\boldsymbol{\theta})$ is a constant-velocity prior on~$\boldsymbol{\theta}$.
The analytical formulation of the likelihood is difficult because there is no observable relation between measurement $\mathbf{x}$ and model parameters $\boldsymbol{\theta}$ available. 
Since our background is static and the objects are textureless, most of the event measurements are generated at the contour of the deforming object. 
We thus assume that each event corresponds to a point on the observed contour of the object. 
We introduce the latent variable~$z_i = j$ which represents the association between the event~$x_i$ and a mesh face of the object with index~$j$. 
 
We use the expectation-maximization (EM) framework to find the model parameters with the latent association,
 \begin{equation}
     \boldsymbol{\theta}^* = \argmax_{\boldsymbol{\theta}} \sum_{i=1}^{N} \ln \sum_{j=1}^{F}  p(x_i , z_i=j \mid \boldsymbol{\theta}) + \ln p(\boldsymbol{\theta}),
 \end{equation}
where~$N$ is the number of aggregated events in an event buffer and~$F$ is the number of mesh faces.
In practice, we aggregate a fixed number of events into an event buffer, assume that the event observations are independent from each other, and optimize over this event buffer. 
The optimization of the next event buffer is initialized with the parameters from the previous buffer.
In the E-step, we update a probabilistic belief on the latent association variables using a variational approximation given the current estimate of parameters~$\overline{\boldsymbol{\theta}}$ from the previous iteration, 
\begin{equation}
    q(z_i) \leftarrow \argmax_{q(z_i)} 
    \sum_{j=1}^{F} q(z_i = j) \ln \frac{p(x_i , z_i=j \mid \overline{\boldsymbol{\theta}})}{q(z_i=j)}.
\end{equation}
The optimal solution of this step is $q(z_i) = p(z_i \mid x_i, \overline{\boldsymbol{\theta}})$. 
In the M-step, the parameters are updated by maximizing the expected log posterior with the probabilistic, i.e. soft, data association from the E-step,
\begin{equation}
    \boldsymbol{\theta} \leftarrow \argmax_{\boldsymbol{\theta}}  \sum_{i=1}^{N} \sum_{j=1}^{F} q(z_i=j) \ln 
    p(x_i , z_i=j \mid \boldsymbol{\theta}) + \ln p(\boldsymbol{\theta}).
\end{equation}
The posterior includes terms for the expected value of the measurement likelihood under the association probability distribution and a prior term on the parameters.
In the following, we explain the concrete form of the EM steps in our approach in detail.

\subsection{Data Association}
Ideally we can use mesh rasterization to find the association of pixels to all mesh faces that intersect the line of sight through each pixel.
Due to limits in the image resolution, initial inaccuracies of the shape parameters during optimization, and complex mesh topologies which allow for multiple layers being intersected by the line of sight, the rasterization often misses the correct association of contour mesh faces with event pixels.
For instance, if the shape estimate is off, the contour mesh face could be observed a few pixels off the actual event location. 
If fingers are bent in front of the palm, events generated on the contour of the finger might hit palm mesh faces, but miss mesh faces on the finger which generate events. 

The EM-framework requires to quantify the probability of associating a mesh face with an event. Intuitively, the closer the mesh face to the event's unprojection ray, the higher the probability it causes the event. Inspired by SoftRasterizer~\cite{SoftRasterizer}, we formulate the contour measurement likelihood as
\begin{equation}
\begin{split}
p(x_i \mid z_i=j, \boldsymbol{\theta})  \propto 
    \sigma \left(\delta^i_j \, \frac{d_{\mathrm{lat}}^2(i, j)}{\alpha}\right)  \exp \left({-\frac{d_{\mathrm{long}}(i, j)}{\beta}}\right)
   \exp & \left({-\frac{r_{\mathrm{ang}}(i, j)}{\gamma}}\right),
\end{split}
\end{equation}
with lateral distance~$d_{\mathrm{lat}}$, longitudinal distance~$d_{\mathrm{long}}$ and angular error $r_{\mathrm{ang}}$ between the line of sight through event~$x_i$ and the mesh face~$f_j$, and sigmoid function $\sigma$.
The angular error~$r_{\mathrm{ang}}$ measures the deviation of the direction of the line of sight from being orthogonal to the normal of the mesh face. 
Hyperparameters~$\alpha$,~$\beta$, and~$\gamma$ are used to control the sharpness of the individual terms for the probability distribution. 

The lateral distance~$d_{\mathrm{lat}}$ is the distance between the line of sight and the closest edge of the mesh face. 
The sign indicator is defined as $\delta_{j}^{i}:=\left\{+1 \text {, if } x_{i} \in f_{j} ;-1 \text {, otherwise }\right\}$. 
We use a maximal lateral distance threshold to reject outlier events due to noise and unmodelled effects.
As the longitudinal distance~$d_\mathrm{long}$, we determine the projected distance between the event pixel and the mesh face center on the line of sight. 
As sketched above, the line of sight may intersect multiple mesh faces on the deformed object. 
The longitudinal distance gives higher likelihood to the mesh face closer to the camera.  
The line of sight through an event caused by the object contour should be approximately orthogonal to the normal of the corresponding mesh face.  
The angular error~$r_{\mathrm{ang}}$ is thus computed by the absolute dot product between the unit direction vector of the line of sight and the face normal.

\subsection{E- and M-Steps}
In the E-step, the association probability is calculated from the measurement likelihood,
\begin{equation}
    p(z_i=j \mid x_i, \boldsymbol{\theta}) = \frac{p(x_i \mid z_i=j, \boldsymbol{\theta}) p(z_i=j \mid \boldsymbol{\theta} )}{\sum_{\substack{j'}} p(x_i \mid z_i=j', \boldsymbol{\theta}) p(z_i=j' \mid \boldsymbol{\theta})} = \frac{p(x_i \mid z_i=j, \boldsymbol{\theta})}{\sum_{\substack{j'}} p(x_i \mid z_i=j', \boldsymbol{\theta})},
\end{equation}
where we assume that the prior probability of the latent variable is uniform. For the M-step, we evaluate the measurement likelihood as
\begin{equation}
\begin{split}
        p(x_i , z_i=j \mid \boldsymbol{\theta}) &= 
    p(x_i \mid z_i=j, \boldsymbol{\theta}) p(z_i=j \mid \boldsymbol{\theta} 
    )
    \propto  \sigma \left(\delta^i_j \,  \frac{d_{\mathrm{lat}}^2(i, j)}{\alpha}\right)
  \exp \left({-\frac{r_{\mathrm{ang}}(i, j)}{\gamma}} \right) .
\end{split}
\end{equation}
Here we do not include the longitudinal distance term.
Ideally, this term should assign a constant probability to mesh faces on the same occlusion layer.
Notably the term does not depend continuously on the shape parameters.
If included in the M-step, our approximative Gaussian term for the E-step would falsely incentivize shape parameters for which the mesh intersects the line of sight closer to camera. 
The angular error term in M-step encourages the alignment of events with contours. 
For scenes with many outlier events (e.g. textured objects), we choose a larger value of~$\gamma$. 
The prior term for the M-step is a constant velocity prior on the parameters, i.e.,
   $\ln p(\boldsymbol{\theta}) =\, k \, \| \mathbf{v} - \mathbf{v}^{'}  \| ^2_2 , ~~~   \mathbf{v} = \frac{\boldsymbol{\theta} - \boldsymbol{\theta}^{'}}{\Delta t}$,
where~$\boldsymbol{\theta}^{'}$ and~$\mathbf{v}^{'}$ are the parameters and velocity for the previous event buffer and~$\Delta t$ is the time difference between the two event buffers. 
We alternate E-step and M-step until convergence. 
When a new event buffer is available, we initialize~$\boldsymbol{\theta}$ based on the current estimate of~$\mathbf{v}$.

\section{Experiments}
We evaluate and demonstrate our event-based non-rigid reconstruction approach on synthetic and real sequences using MANO and SMPL-X object models and involving random motions and various background textures. 
We provide qualitative and quantitative results, comparing with state-of-the-art baselines.
An evaluation of the robustness against noisy events and initial poses, an ablation study for the terms in our E- and M-steps, and results for a hard-EM variant are given in the supplementary material. 
Please also refer to the supplemental video for qualitative results.


\subsection{Experiment Setup}
\label{sec:implementation_details}
\paragraph{Implementation details}
To compensate between the accuracy and the efficiency, we accumulate events in event buffers and optimize a single set of shape parameters for the whole event buffer. 
Similar to \cite{vidal2018_ultimateslam}, we accumulate buffers with a fixed number of events, therefore choosing their temporal length adaptively. 
For our real captured data sequences, we accumulate 100 events per buffer. 
For synthetic data generated by our simulator, we stack 300 events into each buffer. Our simulator simulates the Prophesee camera.
We use the pinhole camera model for the event camera and assume the camera intrinsics are calibrated. 
We have several hyperparameters in our framework. Please refer to the supplementary file for the tuning process of hyperparameters.
Our algorithm optimizes for the pose parameters of the MANO hand model~\cite{mano_hand_model} or the SMPL body model~\cite{smpl2015Loper, smplx}. In case of the hand model, the pose parameters are in PCA space and the MANO modelling approach reconstructs the vertex offsets, which are used together with the canonical pose vertices to generate the posed mesh. For the SMPL model, the pose parameters are the orientation parameters and Linear Blend Skinning (LBS) is used to recover the posed mesh. 

\begin{figure}[tb]
    \centering
    \begin{subfigure}{0.49\textwidth}
        \centering
        \includegraphics[width=0.24\textwidth]{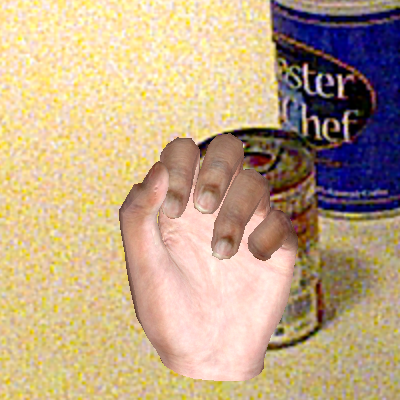}
        \hfill
        \includegraphics[width=0.24\textwidth]{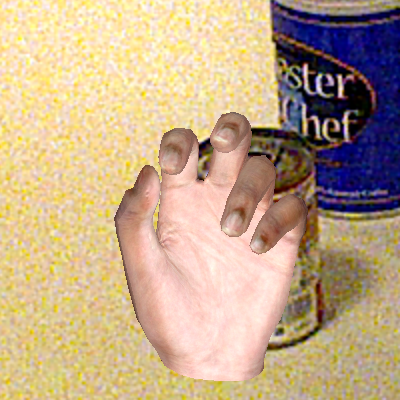}
        \hfill
        \includegraphics[width=0.24\textwidth]{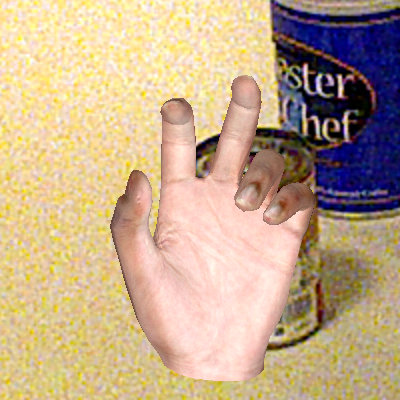}
        \hfill
        \includegraphics[width=0.24\textwidth]{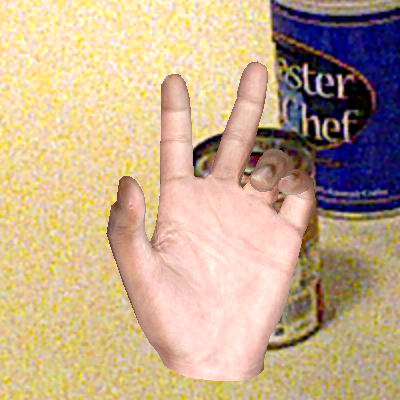}
        \caption{Ground-truth}
        \label{fig:mano_synth_img_gt}
    \end{subfigure}
    \hfill
    \begin{subfigure}{0.49\textwidth}
        \centering
        \includegraphics[width=0.24\textwidth]{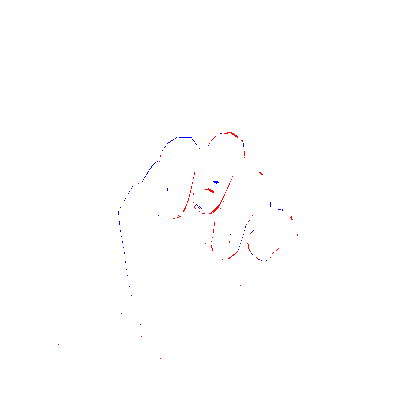}
        \hfill
        \includegraphics[width=0.24\textwidth]{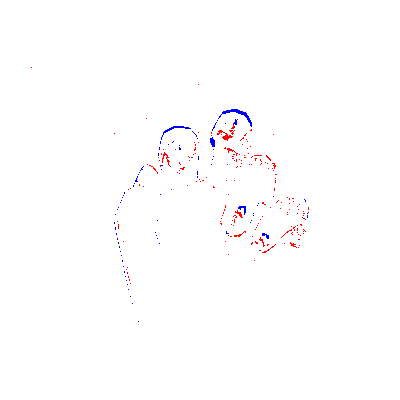}
        \hfill
        \includegraphics[width=0.24\textwidth]{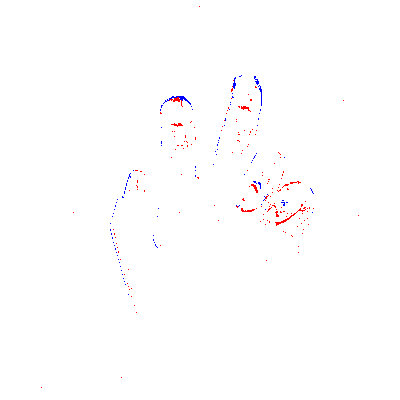}
        \hfill
        \includegraphics[width=0.24\textwidth]{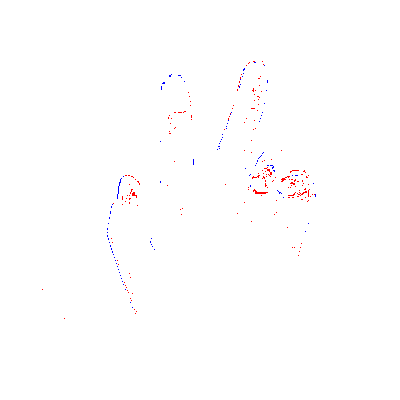}
        \caption{Events}
        \label{fig:mano_synth_events}
    \end{subfigure}
    \hfill
    \begin{subfigure}{0.49\textwidth}
        \centering
        \includegraphics[width=0.24\textwidth]{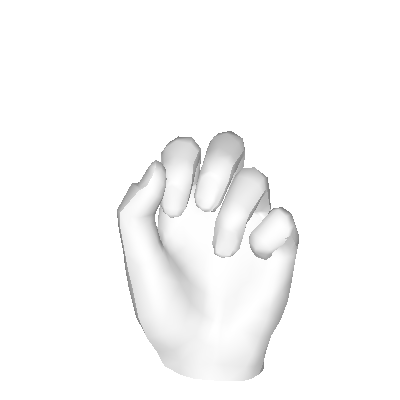}
        \hfill
        \includegraphics[width=0.24\textwidth]{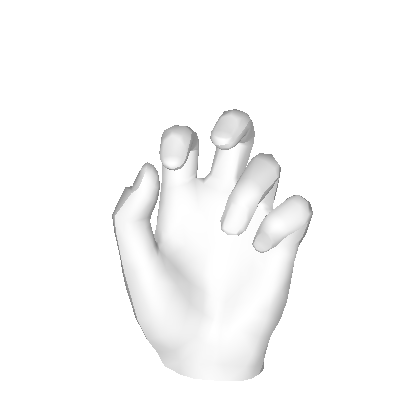}
        \hfill
        \includegraphics[width=0.24\textwidth]{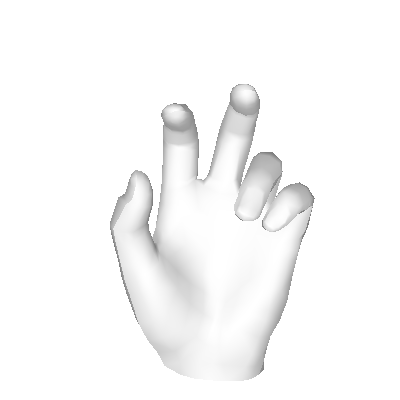}
        \hfill
        \includegraphics[width=0.24\textwidth]{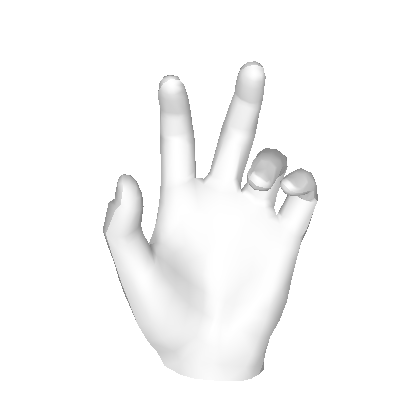}
        \caption{Ours}
        \label{fig:mano_EM_overlaid}
    \end{subfigure}
    \hfill
    \begin{subfigure}{0.49\textwidth}
        \centering
        \includegraphics[width=0.24\textwidth]{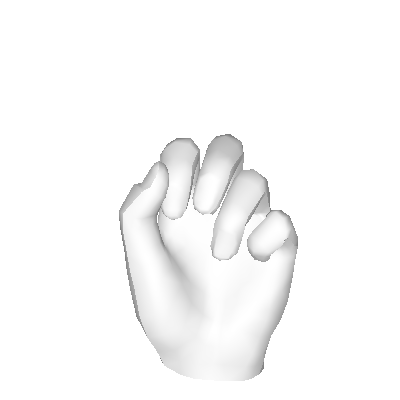}
        \hfill
        \includegraphics[width=0.24\textwidth]{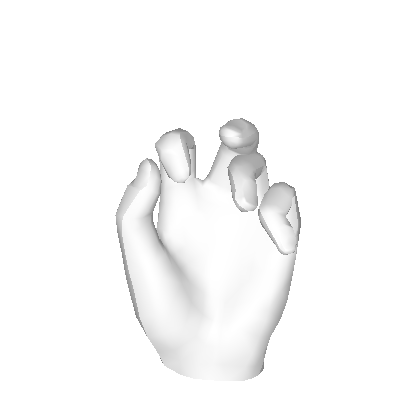}
        \hfill
        \includegraphics[width=0.24\textwidth]{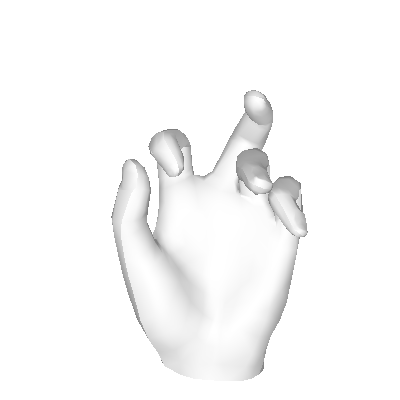}
        \hfill
        \includegraphics[width=0.24\textwidth]{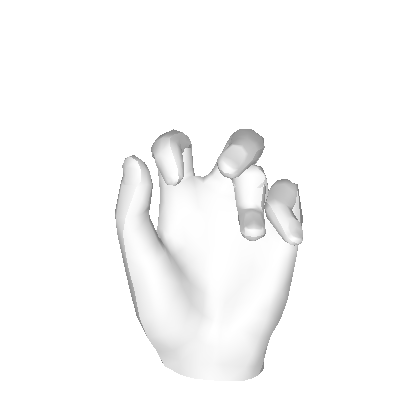}
        \caption{Nehvi's~\cite{nehvi2021_diffevsim}}
        \label{fig:mano_GM_overlaid}
    \end{subfigure}
    \vspace{3mm}
    \caption{Qualitative reconstruction results on synthetic MANO hand sequences.}
    \label{fig:mano_synth}
\end{figure}

\begin{figure}[tb]
    \centering
    \begin{subfigure}{0.49\textwidth}
        \centering
        \includegraphics[width=0.24\textwidth]{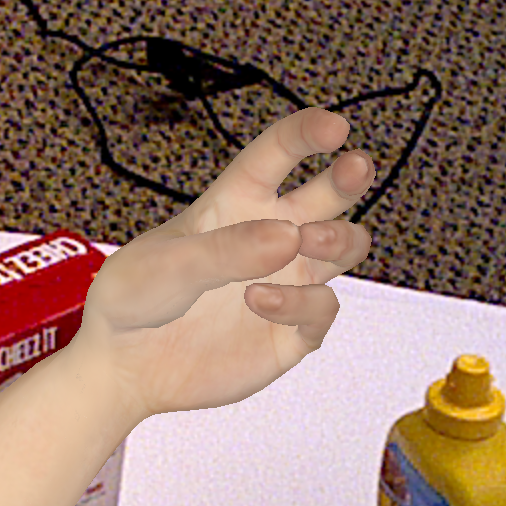}
        \hfill
        \includegraphics[width=0.24\textwidth]{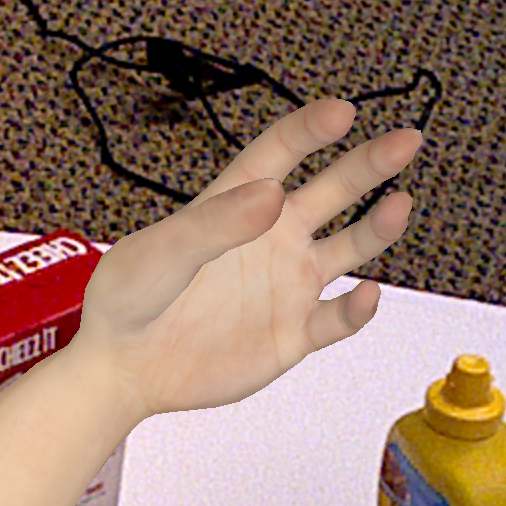}
        \hfill
        \includegraphics[width=0.24\textwidth]{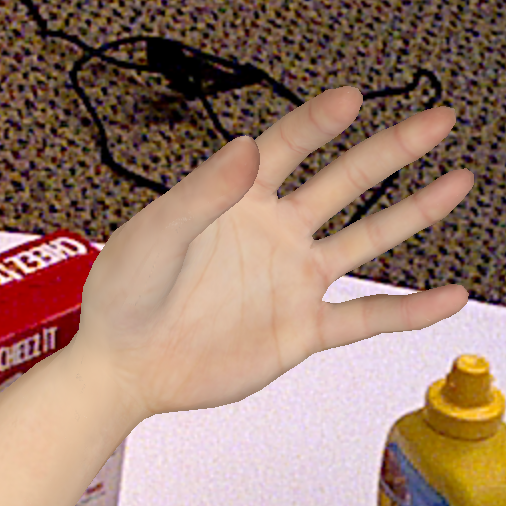}
        \hfill
        \includegraphics[width=0.24\textwidth]{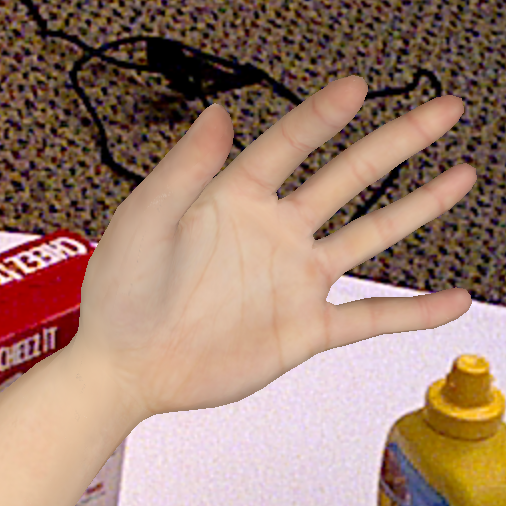}
        \caption{Ground-truth}
        \label{fig:smplx_hand_synth_img_gt}
    \end{subfigure}
    \hfill
    \begin{subfigure}{0.49\textwidth}
        \centering
        \includegraphics[width=0.24\textwidth]{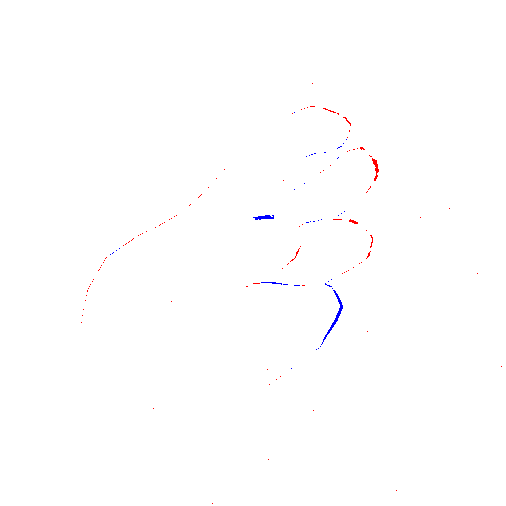}
        \hfill
        \includegraphics[width=0.24\textwidth]{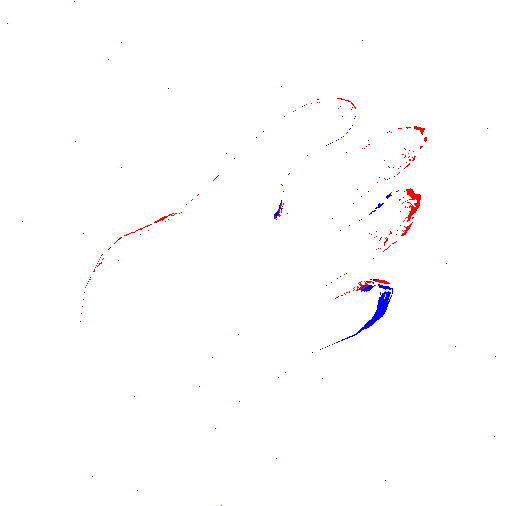}
        \hfill
        \includegraphics[width=0.24\textwidth]{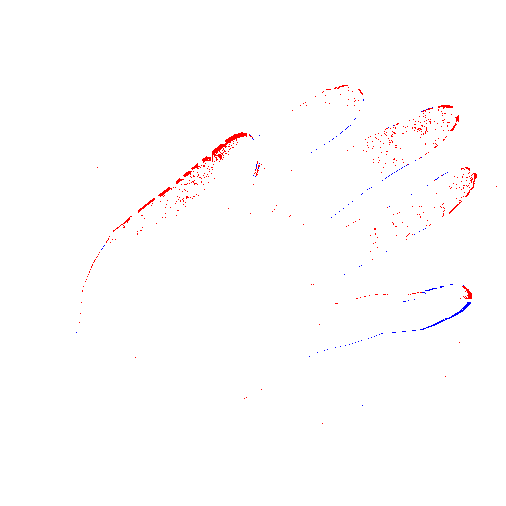}
        \hfill
        \includegraphics[width=0.24\textwidth]{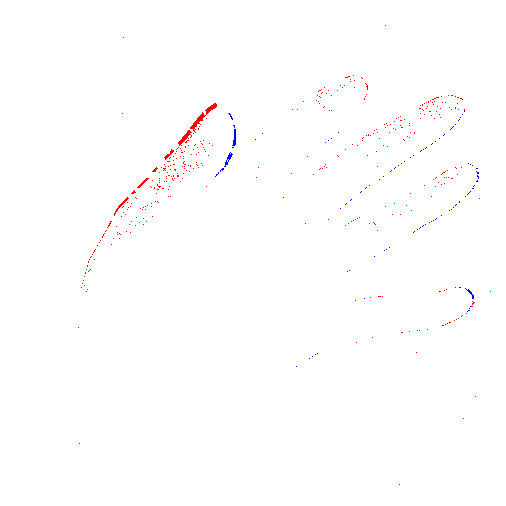}
        \caption{Events}
        \label{fig:smplx_hand_synth_events}
    \end{subfigure}
    \hfill
    \begin{subfigure}{0.49\textwidth}
        \centering
        \includegraphics[width=0.24\textwidth]{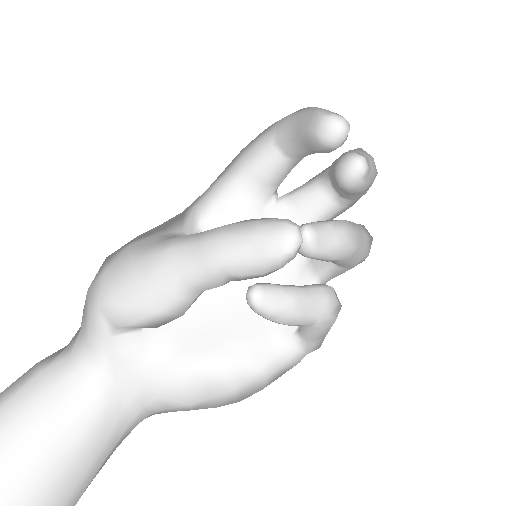}
        \hfill
        \includegraphics[width=0.24\textwidth]{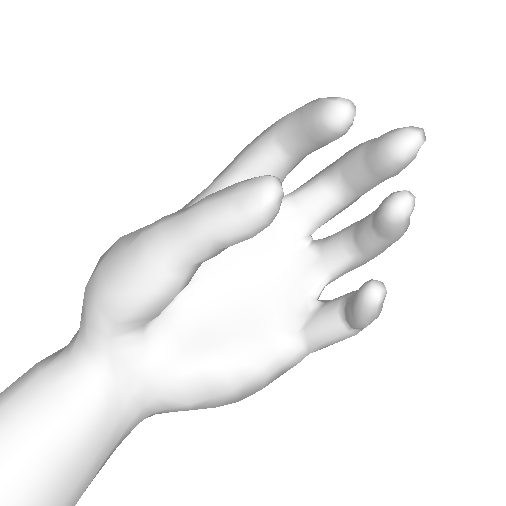}
        \hfill
        \includegraphics[width=0.24\textwidth]{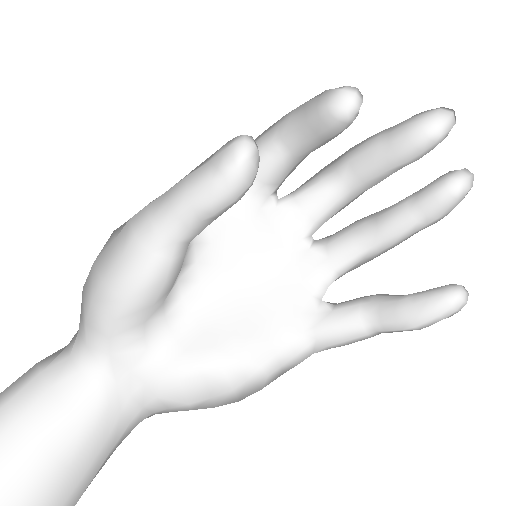}
        \hfill
        \includegraphics[width=0.24\textwidth]{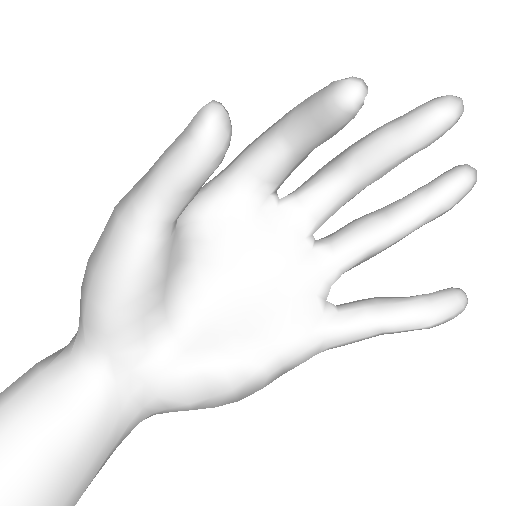}
        \caption{Ours}
        \label{fig:smplx_hand_EM_overlaid}
    \end{subfigure}
    \hfill
    \begin{subfigure}{0.49\textwidth}
        \centering
        \includegraphics[width=0.24\textwidth]{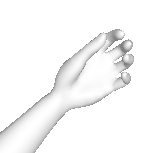}
        \hfill
        \includegraphics[width=0.24\textwidth]{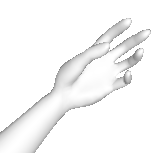}
        \hfill
        \includegraphics[width=0.24\textwidth]{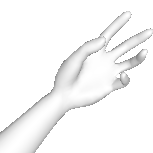}
        \hfill
        \includegraphics[width=0.24\textwidth]{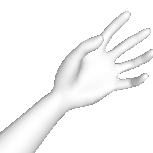}
        \caption{Rudnev's~\cite{rudnev2021_eventhands}}
        \label{fig:smplx_hand_event_hands_overlaid}
    \end{subfigure}
    \vspace{3mm}
    \caption{Qualitative reconstruction results on SMPL-X hand sequences.
    }
    \label{fig:smplxhand_synth}
\end{figure}

\begin{figure}[tb]
    \centering
    \begin{subfigure}{0.49\textwidth}
        \centering
        \includegraphics[width=0.24\textwidth]{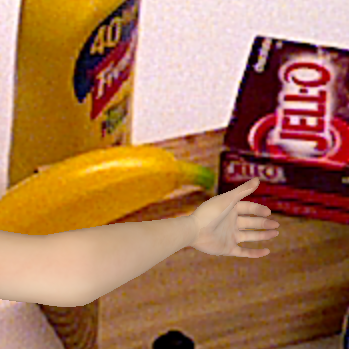}
        \hfill
        \includegraphics[width=0.24\textwidth]{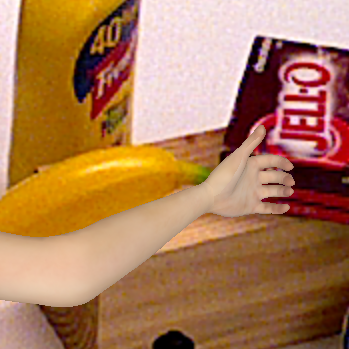}
        \hfill
        \includegraphics[width=0.24\textwidth]{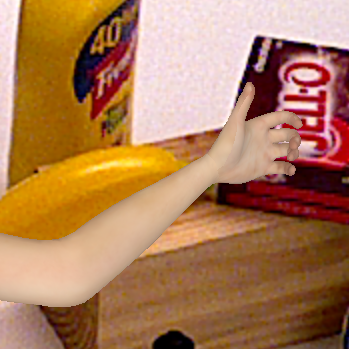}
        \hfill
        \includegraphics[width=0.24\textwidth]{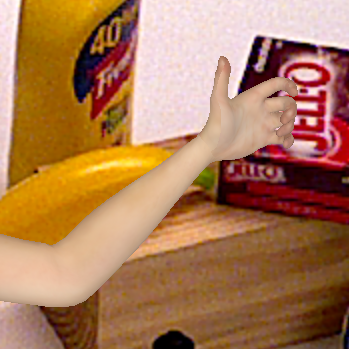}
        \caption{Ground-truth}
        \label{fig:smplx_body_hand_synth_img_gt}
    \end{subfigure}
    \hfill
    \begin{subfigure}{0.49\textwidth}
        \centering
        \includegraphics[width=0.24\textwidth]{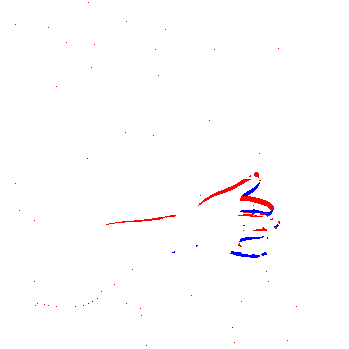}
        \hfill
        \includegraphics[width=0.24\textwidth]{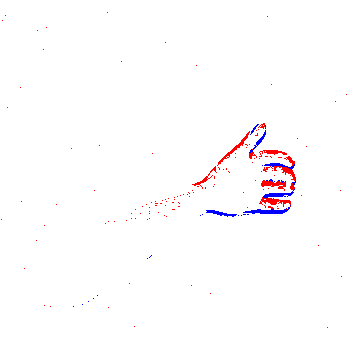}
        \hfill
        \includegraphics[width=0.24\textwidth]{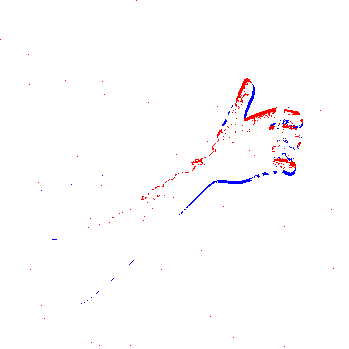}
        \hfill
        \includegraphics[width=0.24\textwidth]{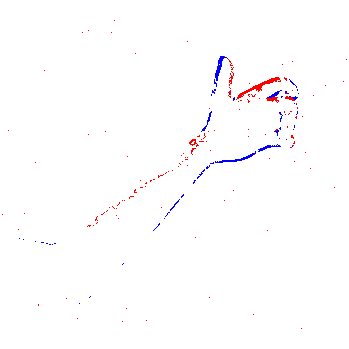}
        \caption{Events}
        \label{fig:smplx_body_hand_synth_events}
    \end{subfigure}
    \hfill
    \begin{subfigure}{0.49\textwidth}
        \centering
        \includegraphics[width=0.24\textwidth]{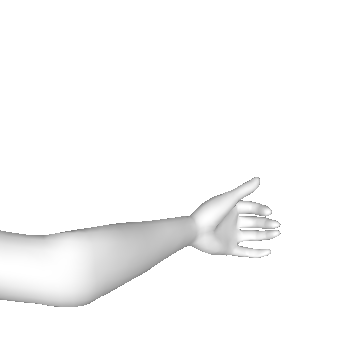}
        \hfill
        \includegraphics[width=0.24\textwidth]{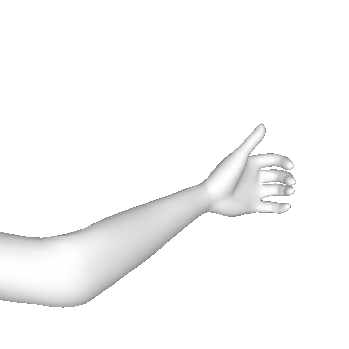}
        \hfill
        \includegraphics[width=0.24\textwidth]{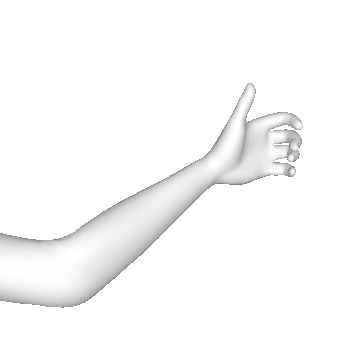}
        \hfill
        \includegraphics[width=0.24\textwidth]{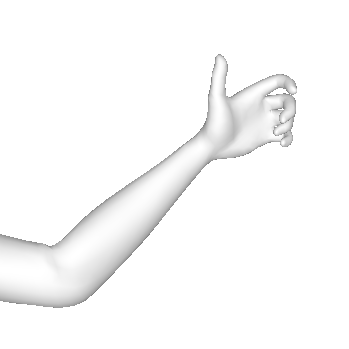}
        \caption{Ours}
        \label{fig:smplx_body_hand_EM_overlaid}
    \end{subfigure}
    \vspace{3mm}
    \caption{Qualitative reconstruction results on SMPL-X arm and hand sequences.}
    \label{fig:smplxarm_synth}
\end{figure}

\paragraph{Datasets}
We generate synthetic datasets of sequences with three types of different objects, namely the MANO~\cite{mano_hand_model} hand~(Fig. \ref{fig:mano_synth_img_gt}), the SMPL-X~\cite{smpl2015Loper, smplx} hand~(Fig. \ref{fig:smplx_hand_synth_img_gt}), and the combined SMPL-X~\cite{smpl2015Loper, smplx} arm and hand~(\ref{fig:smplx_body_hand_synth_img_gt}). 
MANO hand sequences are generated with a single hand mesh at a fixed position and orientation. 
We vary the full 45-dimensional pose parameter space to generate varying hand poses. 
For SMPL-X hand sequences, the hand is attached to the whole human body which prevents observing the inside of the hand mesh. 
We vary the first 6 principal component pose parameters to simulate time-varying and realistic hand deformations. 
In the SMPL-X arm and hand sequences, we synthesize the arm motion by the 3-DoF rotation of the elbow joint and the hand motion by the 6 principal pose parameters.
We use a custom event simulator with adaptive sampling rate to generate the synthetic sequences for the different object models.
A detailed description of the event simulator can be found in the supplementary material.
We simulate a Prophesee camera with the image size of~$1280\times720$. 
The event contrast threshold is~$0.5$. 
For each sequence, the background image is randomly chosen from a texture-rich indoor scene in the YCB video dataset~\cite{yu_posecnn_rss18}. 
To introduce noise into the event generation process, we sample the contrast threshold of each pixel from a Gaussian distribution with standard deviation~$0.0004$. 
The threshold of salt-and-pepper noise is~$10^{-5}$. 
For further details on the noise generation process in our simulator, please refer to the supplementary material.

\paragraph{Evaluation metrics}
For the synthetic data, 3D ground-truth positions for all joints and mesh vertices as well as pose parameters are known. 
We evaluate using the Mean Per Joint Position Error (MPJPE~\cite{vonMarcard2018_3dpw}), the percentage of correct 3D Joints (3D-PCK~\cite{dushyant2017_3dhp}), and the area under the PCK-curve (AUC~\cite{dushyant2017_3dhp}) with thresholds ranging from 0 to 50\,mm. 
For hand sequences, we consider the 15 hand skeleton joints. For arm and hand sequences, only the forearm and the hand have motion. Thus, we consider one wrist joint and 15 hand joints.

\subsection{Quantitative Evaluation}
\label{sec:quantitative}

We compare our approach quantitatively with the state-of-the-art event-based non-rigid object reconstruction methods: Nehvi's optimization-based approach \cite{nehvi2021_diffevsim} evaluates using the MANO hand model, while Rudnev's approach \cite{rudnev2021_eventhands} is designed for the SMPL-X hand model. 
To the best of our knowledge, previous event-based reconstruction approaches have not been demonstrated on combined arm and hand motion of a SMPL-X model.
Hence, we only provide results for our method on these sequences.

\begin{table}[tb]
    \centering
    \begin{tabular}{lccc}
    \hline 
    Scenario        & Method & mean MPJPE $(mm)$ & median MPJPE $(mm)$ \\
    \hline 
    \multirow{2}{*}{MANO hand} & Nehvi et al. \cite{nehvi2021_diffevsim} & $11.61$ & $10.85$ \\
    
                & Ours & $\textbf{4.52}$ & $\textbf{4.27}$  \\
   \hline
   \multirow{2}{*}{SMPL-X hand} & Rudnev et al. \cite{rudnev2021_eventhands} & $11.88$ & $10.73$\\
                & Ours & $\textbf{1.11}$ & $\textbf{0.76}$\\
    \hline
     SMPL-X arm \& hand & Ours & $\textbf{15.39}$ & $\textbf{3.93}$\\
     \hline
    \end{tabular}
    \vspace{1mm}
    \caption{Results on synthetic sequences of different objects}
    \label{tab:mano_synth_quantitative}
\end{table}

\begin{figure}[tb]
\centering

    \begin{subfigure}{0.33\textwidth}
        \centering
        \includegraphics[width=\textwidth]{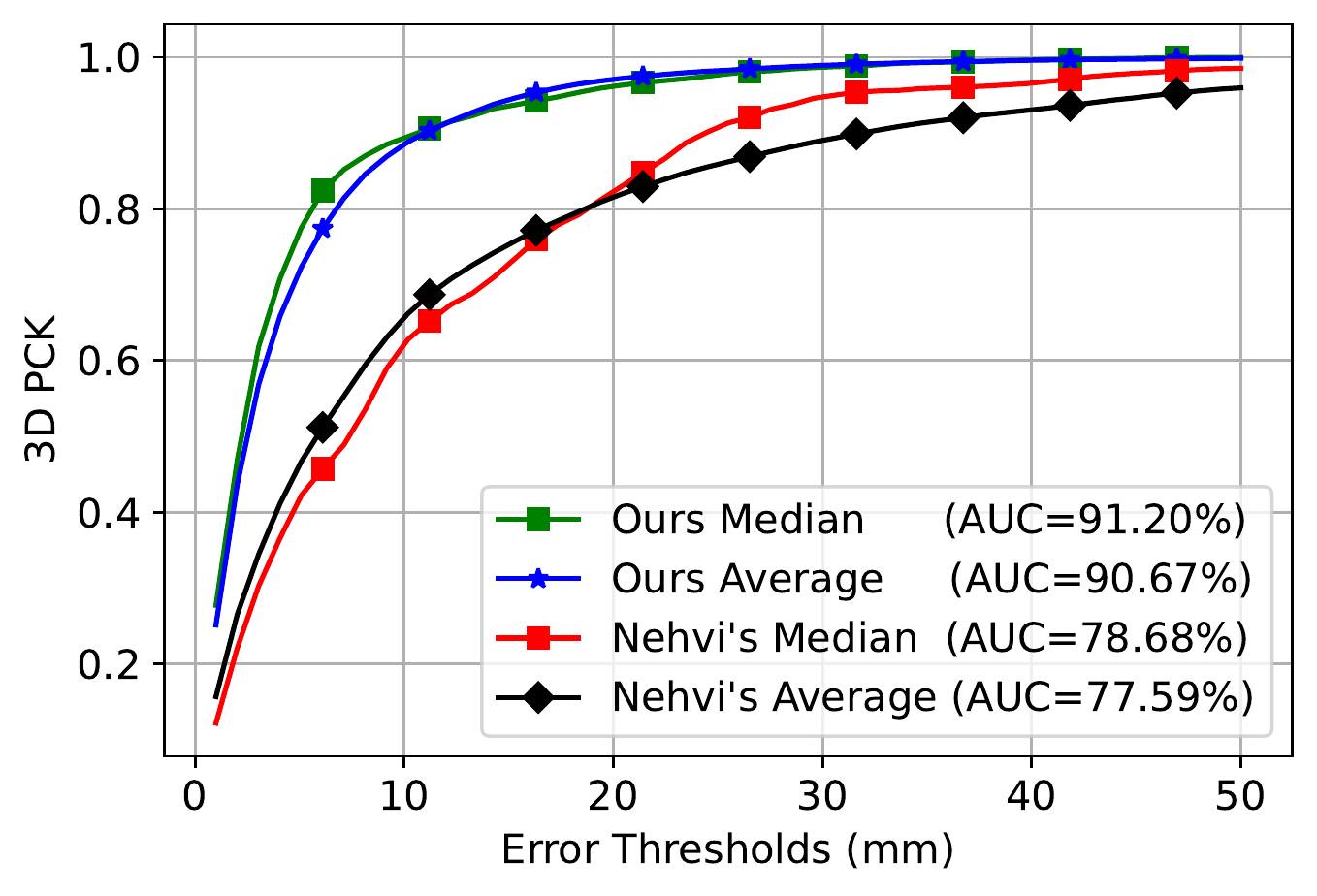}
        \caption{MANO hand sequences.}
        \label{fig:3d_pck_mano_synth}
    \end{subfigure}
    \begin{subfigure}{0.32\textwidth}
        \centering
        \includegraphics[width=\textwidth]{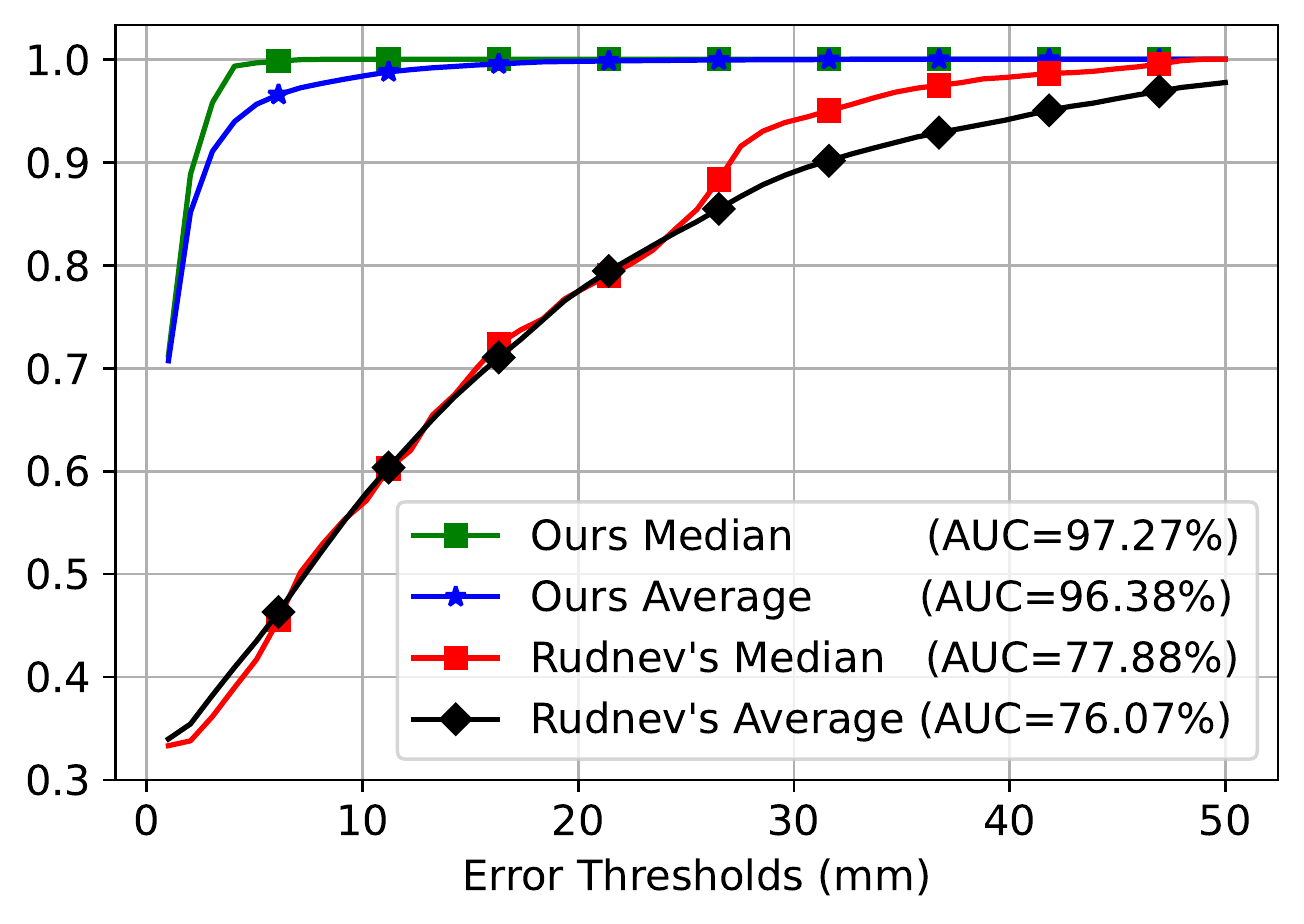}
        \caption{SMPL-X hand sequences.}
        \label{fig:3d_pck_smplx_hand}
    \end{subfigure}
    \begin{subfigure}{0.32\textwidth}
        \centering
        \includegraphics[width=\textwidth]{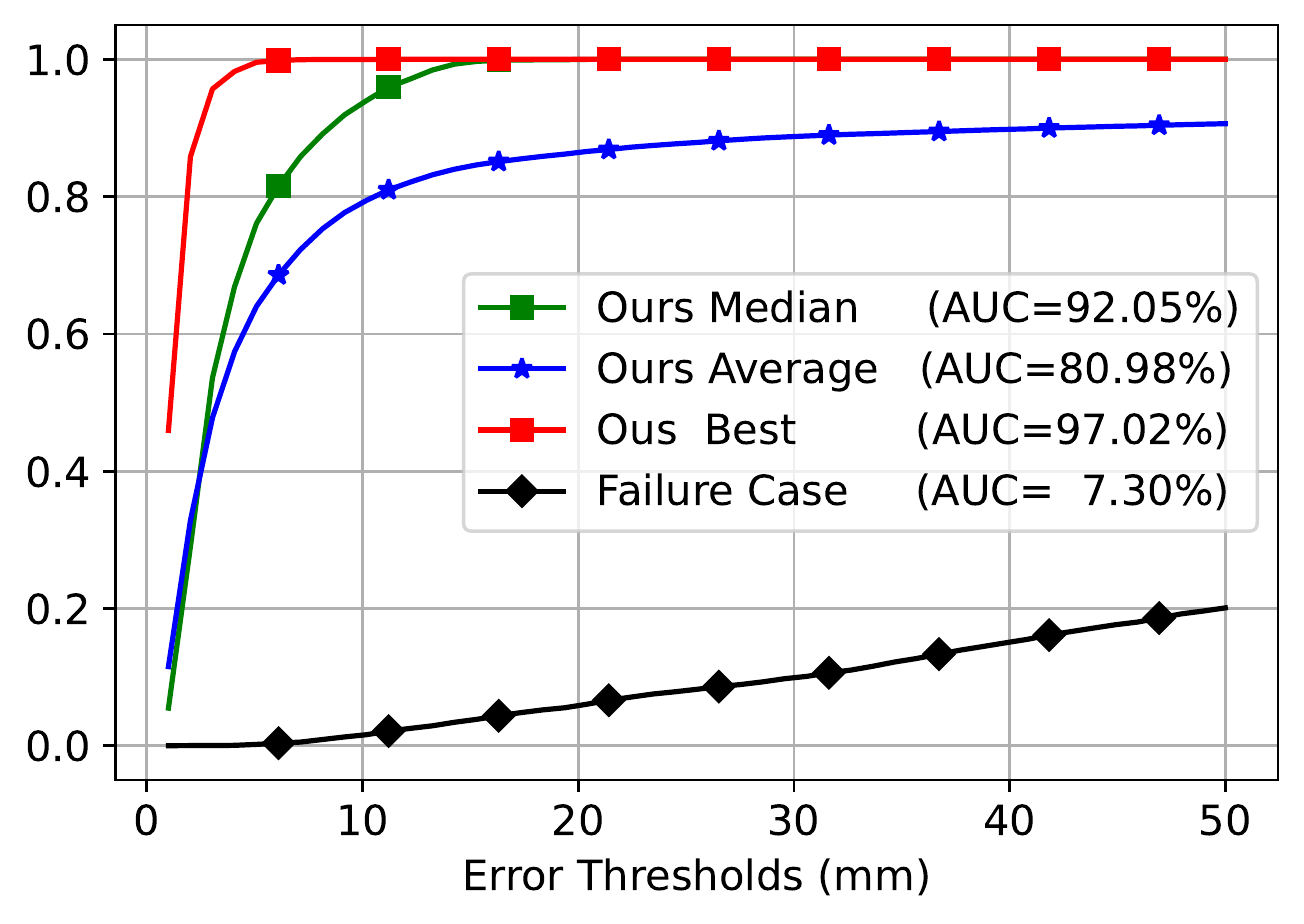}
        \caption{SMPL-X arm+hand seqs.}
        \label{fig:3d_pck_arm_hand}
    \end{subfigure}
    \vspace{3mm}
    \caption{3D-PCK curve $@50mm$ on synthetic sequences.}
\end{figure}


For synthetic MANO hand sequences, we use the MANO~\cite{mano_hand_model} hand model as the parametric mesh template. 
In experiments, we initialize the optimized parameters with the ground-truth pose parameters and evaluate, how well the approach can keep track of the hand deformation. 
Our approach reconstructs the 45-dimensional pose parameter. 
We report quantitative results MPJPE and AUC on these sequences in Table~\ref{tab:mano_synth_quantitative}. 
We also compare our method to the state-of-the-art event-based hand tracking approach~\cite{nehvi2021_diffevsim} for the MANO model. 
Similar to our approach, Nehvi's method is optimization-based and requires the initial parameters of the mesh template. To ensure a fair comparison, we use Optuna~\cite{optuna} to tune hyperparameters in Nehvi's method and our method.
We observe that our approach is about $2.5$-times more accurate than Nehvi's method. 
We also show the 3D-PCK curve of both approaches in Fig.~\ref{fig:3d_pck_mano_synth}. 
Apparently, our method has higher AUC than Nehvi's method. 
Results in Table~\ref{tab:mano_synth_quantitative} and Fig.~\ref{fig:3d_pck_mano_synth} demonstrate that our method outperforms Nehvi's method clearly.

In the SMPL-X hand sequences, the hand is attached to a human body model. 
Here, our approach reconstructs the 6-dimensional pose parameters, which is consistent with the evaluation conducted in Rudnev's method~\cite{rudnev2021_eventhands}. 
We report quantitative results in Table~\ref{tab:mano_synth_quantitative}. 
It can be seen that our approach achieves better performance than Rudnev's method~\cite{rudnev2021_eventhands} in MPJPE. 
Rudnev's method is learning-based and does not require the knowledge of the initial pose parameters. 
We use the network trained by~\cite{rudnev2021_eventhands} which is limited to the resolution $(240\times180)$ of the DAVIS 240C camera. 
Thus, we simulate event streams of the same motion with the intrinsics provided in~\cite{rudnev2021_eventhands} for Rudnev's method. 
We observe that since the global rotation and translation are fixed, events are only generated where the deformation occurs. 
Rudnev's method seems to perform less well in this case than our approach. 

Finally, we evaluate the performance of our approach on sequences which combine arm and hand motion using the SMPL-X model. 
In the synthetic data generation process, we vary 6 principal parameters to synthesize hand poses and the 3 rotation parameters of the elbow joint. 
Our approach jointly optimize these hand and elbow parameters. 
The median MPJPE in Tab.~\ref{tab:mano_synth_quantitative} demonstrates that our approach can reconstruct the motion of the arm and hand with high accuracy. 
The mean MPJPE is higher than the median MPJPE due to failures in some sequences.
We show failure cases and their analysis in the supplementary material.
The difference in accuracy to the SMPL-X hand sequences can be explained by the fact that for the SMPL-X arm \& hand sequences, also the elbow joint needs to be reconstructed.
Moreover, the hand is visible on different scales in the image (the hand is smaller for SMPL-X arm \& hand).
Hence, the absolute error in \,mm becomes higher. 
As an incremental optimization-based approach, our approach can also drift, but it can snap the mesh silhouette to the observed events on the contour if sufficient observations are available. In the supplementary material, we provide a plot of median error over time for the MANO hand dataset. 

Differently to~\cite{meshGraphormer,rudnev2021_eventhands}, our implementation is not real-time capable, due to the complete evaluation of all event measurement likelihoods for all mesh faces in the soft E-step.  
Depending on the motion, a sequence can be split into 50 to 500 event buffers. For each event buffer, the current average run-time of the method is $8.76$ seconds on MANO hand sequences, and $50.72$ seconds on SMPL-X arm and hand sequences.
Our current implementation in PyTorch. In future work, the optimization process could be implemented more efficiently by associating mesh faces with a local search, tailor code with CUDA/C++, and using second-order Gauss-Newton methods instead of the current gradient-descent algorithm.
In the supplementary material, we also provide evidence that hard data association with the most likely mesh face in the E-step only slightly affects accuracy.
Due to the non-convexity of the problem, our approach needs a sufficiently good initial guess of the pose. 
In the supplementary material, we evaluate reconstruction accuracy vs. varying noise levels for the initial pose.
We also evaluate the effect of varying noise in the events on accuracy in the supplementary material.


\subsection{Qualitative Evaluation}
\label{sec:qualitative}

\paragraph{Synthetic Data}
We show qualitative results of our approach and state-of-the-art baseline approaches~\cite{nehvi2021_diffevsim, rudnev2021_eventhands} on synthetic sequences. 
For each object, the ground-truth RGB images, accumulated events during the motion, and reconstruction results are shown. 
We crop all images with a fixed ratio to increase the view of the objects.
Results on synthetic MANO hand sequences of our approach and Nehvi's approach~\cite{nehvi2021_diffevsim} are shown in Figs.~\ref{fig:mano_EM_overlaid} and~\ref{fig:mano_GM_overlaid}, respectively. 
It can be observed that our approach reconstructs the deformation of the hand well, while Nehvi's approach struggles to track the hand pose accurately.
Note that Nehvis method does assume black background and generatively models the specific log intensity changes induced at the optical flow at contours.
Our approach only assumes that events are generated by contours without explicit dependency on the optical flow, hence, it is more robust to textured backgrounds.
We compare our approach with Rudnev's method~\cite{rudnev2021_eventhands} in Fig.~\ref{fig:smplx_hand_event_hands_overlaid} on a SMPL-X hand deformation sequence. 
While our approach can reconstruct the hand motion well, Rudnev's approach performs less accurately.
For the sequences with combined arm and hand motion of the SMPL-X model, we show qualitative results of our approach in Fig.~\ref{fig:smplx_body_hand_EM_overlaid}. 
Our proposed approach can reconstruct the motion well.

\begin{figure}[!t]
\setlength\tabcolsep{2pt}%
\begin{tabularx}{\textwidth}{>{\centering\arraybackslash}m{1.2cm} >{\centering\arraybackslash}m{1.51cm} >{\centering\arraybackslash}m{1.51cm} >{\centering\arraybackslash}m{1.51cm} >{\centering\arraybackslash}m{1.51cm} >{\centering\arraybackslash}m{1.51cm} >{\centering\arraybackslash}m{1.51cm} >{\centering\arraybackslash}m{1.51cm}}
    Image (24 fps)
    &
   \includegraphics[ width=1\linewidth, height=1\linewidth, keepaspectratio]{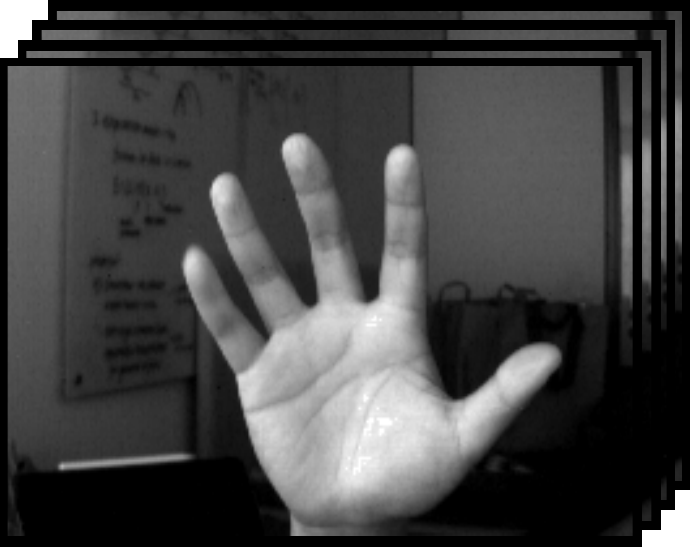} 
   &
   &
   \includegraphics[ width=1\linewidth, height=1\linewidth, keepaspectratio]{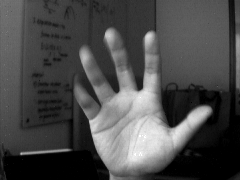} 
   &
   &
   \includegraphics[ width=1\linewidth, height=1\linewidth, keepaspectratio]{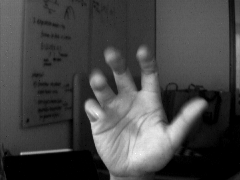} 
   & 
   &
   \includegraphics[ width=1\linewidth, height=1\linewidth, keepaspectratio]{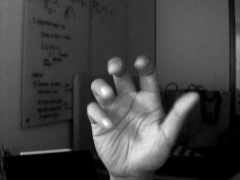} \\
   
   Lin et al. \cite{meshGraphormer}
   &
   \includegraphics[ width=1\linewidth, height=1\linewidth, keepaspectratio]{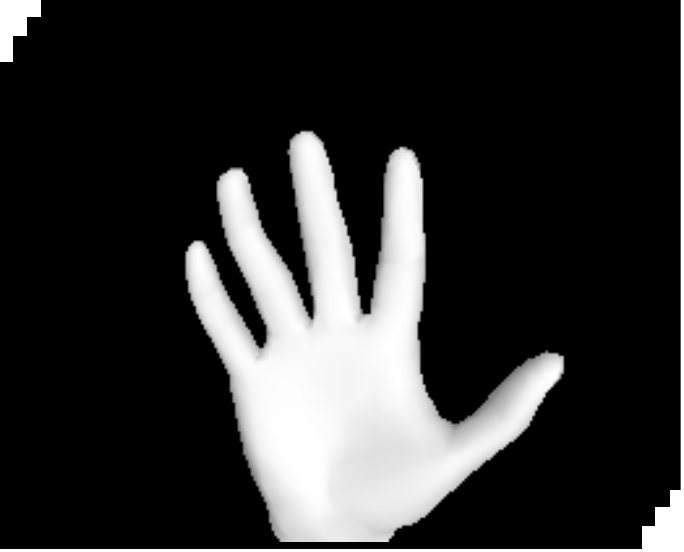} 
   &
   &
   \includegraphics[ width=1\linewidth, height=1\linewidth, keepaspectratio]{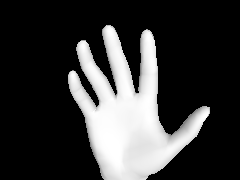}
   &
   &
   \includegraphics[ width=1\linewidth, height=1\linewidth, keepaspectratio]{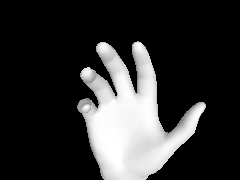} 
   & 
   &
   \includegraphics[ width=1\linewidth, height=1\linewidth, keepaspectratio]{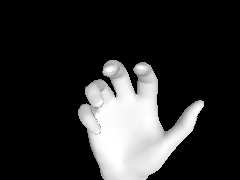} \\
   
   Events
   &
   \includegraphics[ width=1\linewidth, height=1\linewidth, keepaspectratio]{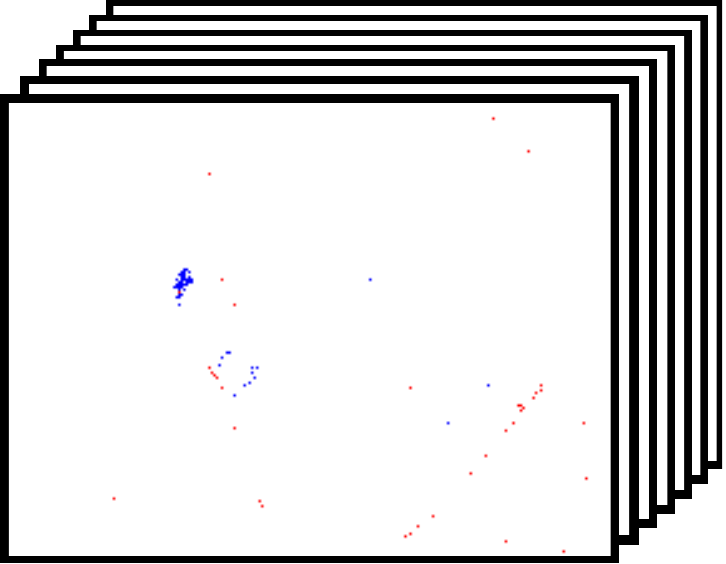} 
   &
   \includegraphics[ width=1\linewidth, height=1\linewidth, keepaspectratio]{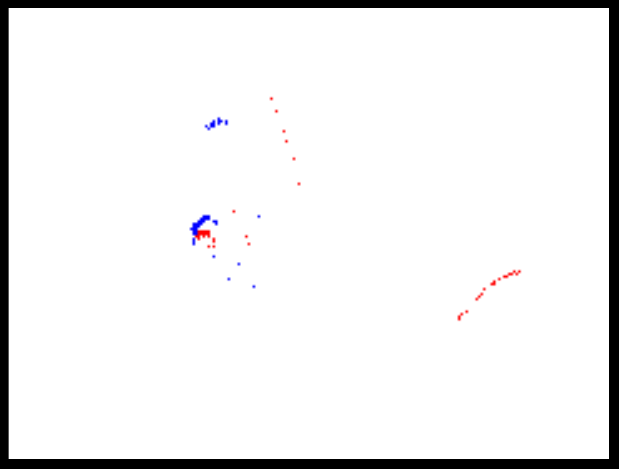} 
   &
   \includegraphics[ width=1\linewidth, height=1\linewidth, keepaspectratio]{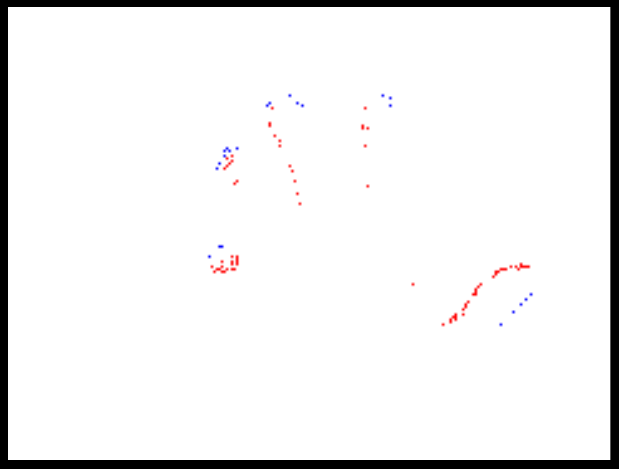} 
   &
   \includegraphics[ width=1\linewidth, height=1\linewidth, keepaspectratio]{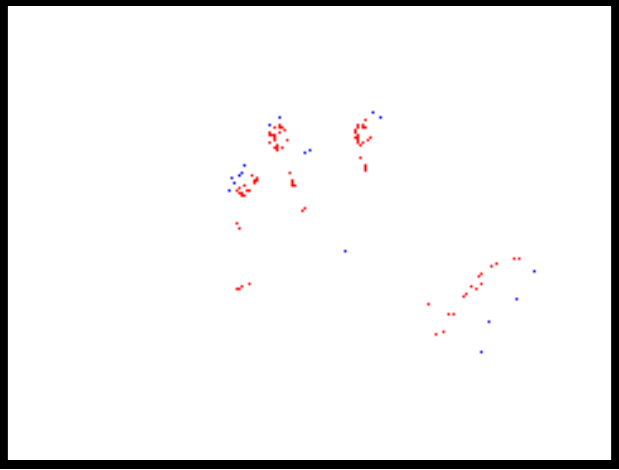}
   &
   \includegraphics[ width=1\linewidth, height=1\linewidth, keepaspectratio]{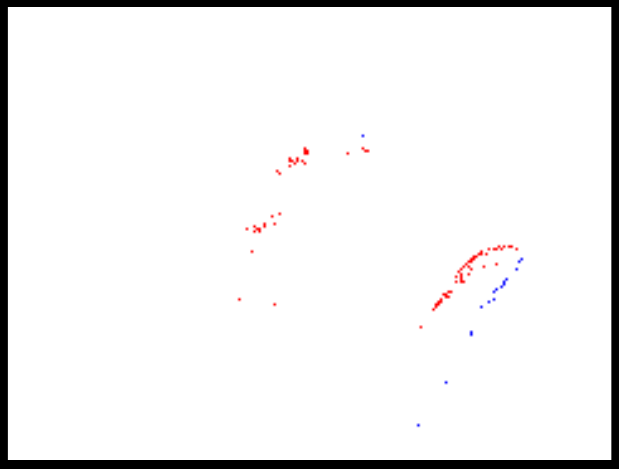}
   &
   \includegraphics[ width=1\linewidth, height=1\linewidth, keepaspectratio]{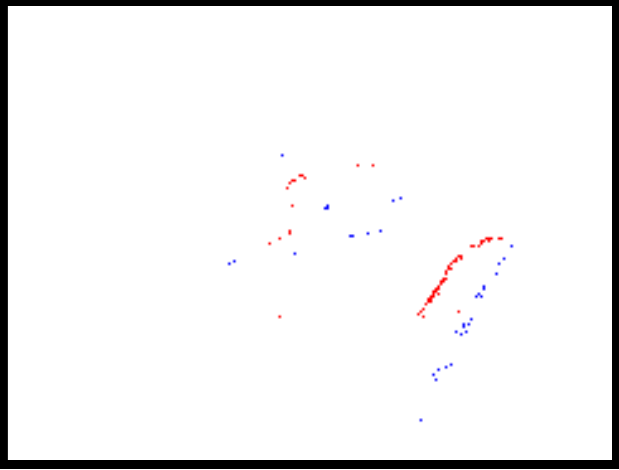}
   &
   \includegraphics[ width=1\linewidth, height=1\linewidth, keepaspectratio]{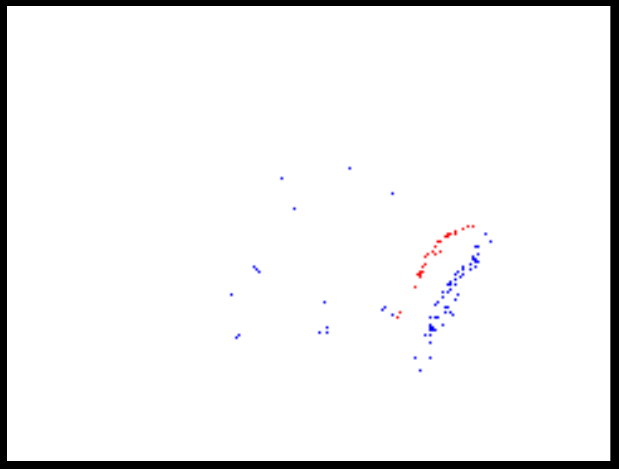} \\
      
   Nehvi et al. \cite{nehvi2021_diffevsim}
   &
   \includegraphics[ width=1\linewidth, height=1\linewidth, keepaspectratio]{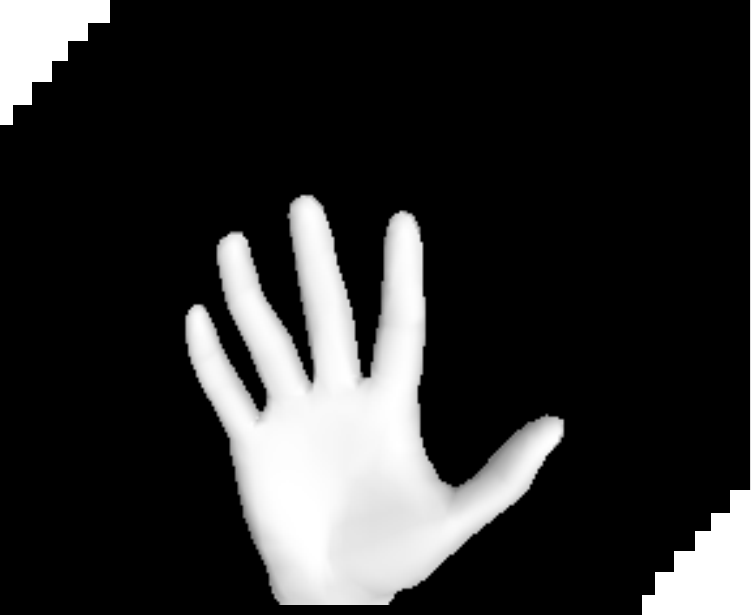} 
   &
   \includegraphics[ width=1\linewidth, height=1\linewidth, keepaspectratio]{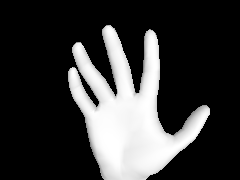} 
   &
   \includegraphics[ width=1\linewidth, height=1\linewidth, keepaspectratio]{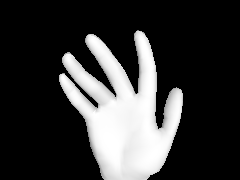} 
   &
   \includegraphics[ width=1\linewidth, height=1\linewidth, keepaspectratio]{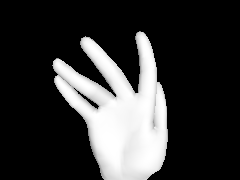} 
   &
   \includegraphics[ width=1\linewidth, height=1\linewidth, keepaspectratio]{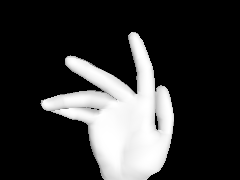} 
   &
   \includegraphics[ width=1\linewidth, height=1\linewidth, keepaspectratio]{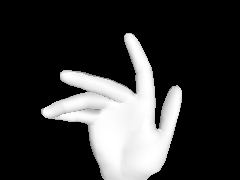} 
   &
   \includegraphics[ width=1\linewidth, height=1\linewidth, keepaspectratio]{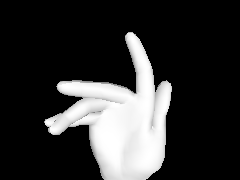} \\
   
  Rudnev et al. \cite{rudnev2021_eventhands}
   &
   \includegraphics[ width=1\linewidth, height=1\linewidth, keepaspectratio]{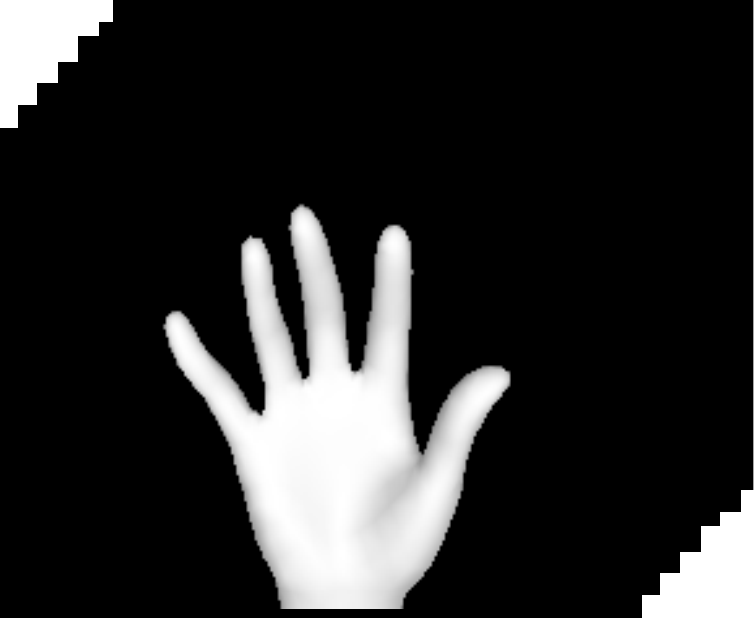} 
   &
   \includegraphics[ width=1\linewidth, height=1\linewidth, keepaspectratio]{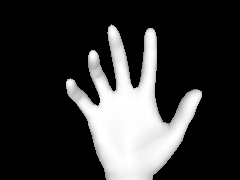} 
   &
   \includegraphics[ width=1\linewidth, height=1\linewidth, keepaspectratio]{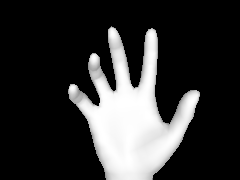} 
   &
   \includegraphics[ width=1\linewidth, height=1\linewidth, keepaspectratio]{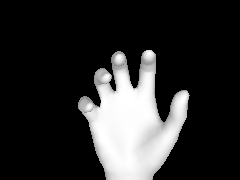} 
   &
   \includegraphics[ width=1\linewidth, height=1\linewidth, keepaspectratio]{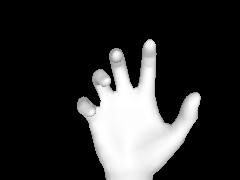} 
   &
   \includegraphics[ width=1\linewidth, height=1\linewidth, keepaspectratio]{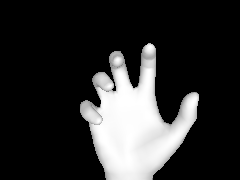} 
   &
   \includegraphics[ width=1\linewidth, height=1\linewidth, keepaspectratio]{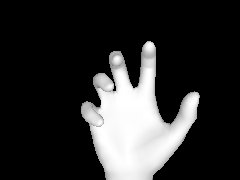} \\

   Ours
   & 
   \includegraphics[ width=1\linewidth, height=1\linewidth, keepaspectratio]{images/stacked_gm_em.pdf}
   &
   \includegraphics[ width=1\linewidth, height=1\linewidth, keepaspectratio]{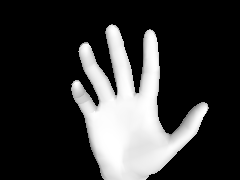} 
   &
   \includegraphics[ width=1\linewidth, height=1\linewidth, keepaspectratio]{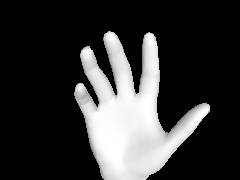}
   &
   \includegraphics[ width=1\linewidth, height=1\linewidth, keepaspectratio]{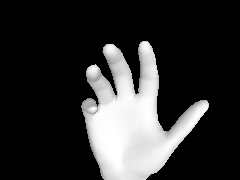} 
   &
   \includegraphics[ width=1\linewidth, height=1\linewidth, keepaspectratio]{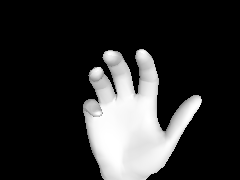} 
   &
   \includegraphics[ width=1\linewidth, height=1\linewidth, keepaspectratio]{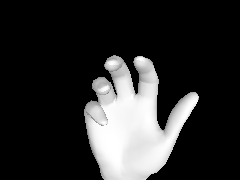}
   &
   \includegraphics[ width=1\linewidth, height=1\linewidth, keepaspectratio]{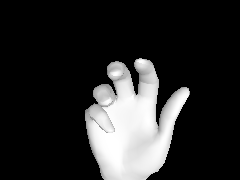} \\

\end{tabularx}
\caption{Qualitative results on a real event sequence from a DAVIS240C camera. Lin et al.~\cite{meshGraphormer} infer MANO pose parameters from intensity images; Rudnev et al.~\cite{rudnev2021_eventhands} infer 6 principal MANO pose parameters; Nehvi et al.~\cite{nehvi2021_diffevsim} and our approach optimize 45 MANO pose parameters. Our approach recovers the deformation most similar to the ground truth. } 
\label{fig:qualitative_davis}
\end{figure}

\paragraph{Real Data}
In Fig.~\ref{fig:qualitative_davis}, we also show qualitative results of our approach with the MANO hand model on real sequences with hand motion captured with a DAVIS240C camera. 
The camera also records grayscale intensity frames for reference. 
Since our approach requires an initialization of the hand pose parameters, we use~\cite{meshGraphormer} on the first image frame and set rotation and translation manually, since the pretrained model did not yield proper poses on the DAVIS gray scale images.
Further details on the initialization procedure are provided in the supplementary material.
We compare our approach qualitatively with state-of-the-art image-based (MeshGraphormer~\cite{meshGraphormer}) and event-based~\cite{nehvi2021_diffevsim, rudnev2021_eventhands} methods. 
MeshGraphormer is a learning-based approach which predicts MANO pose parameters from grayscale images. 
It has solid reconstruction performance for slower motions, but suffers from motion blur for fast motions. 
Furthermore, the temporal resolution of the reconstruction result is limited by the frequency of the frames. 
Compared to the result of MeshGraphormer~\cite{meshGraphormer} and event-based approaches~\cite{nehvi2021_diffevsim, rudnev2021_eventhands}, our approach follows the ground-truth reference more closely. 

\subsection{Assumptions and Limitations}

Our approach uses a loose coupling of frames and events by initializing the optimization from the gray-scale frame. 
A possible direction of future work is to extend the method by feeding the frame-based information at a specific lower rate and use the events to estimate pose between frames in a tightly-coupled joint optimization framework. 
In our experiments, self-occlusions occur within the hand (for instance between fingers, or fingers and the palm, see also Fig.~\ref{fig:mano_synth}). 
The more self occlusions, the more unconstrained the pose parameters get due to the partial observations and low number of events. 
The constant velocity model and the PCA subspace of MANO can help to regularize the motion in this partially constrained setting. 
Our method relies on events on the contour and cannot estimate deformation if there are little contour events due to similar background color or insufficient motion.
Due to the image projection, the contour information seems not sufficient yet for reconstructing shape parameters concurrently with rotation and translation of the objects with our formulation. 
To address challenging settings like 6D pose estimation or crossing hands in future work, one could for instance investigate including learned temporal priors, texture-based cues, or combining events with frames in a joint optimization framework.

\section{Conclusion}
We present a novel non-rigid reconstruction approach for event cameras.
Our approach formulates the reconstruction problem as an expectation-maximization problem.
Events are associated to observed contours on parametrized mesh models and an alignment objective is maximized to fit the mesh parameters with event measurements.
Our method outperforms qualitatively and quantitatively state-of-the-art event-based non-rigid reconstruction approaches~\cite{nehvi2021_diffevsim, rudnev2021_eventhands}. 
We also demonstrate that our proposed approach is robust to noisy events and initial parameter estimates. 
In future work, texture-based reconstruction from events and frames could be combined with our approach or the run-time of our implementation could be improved by searching for correspondences efficiently.


\section*{Acknowledgement}
This work was supported by Cyber Valley and the Max Planck Society. The authors thank the Empirical Inference Department at Max Planck Institute for Intelligent Systems for providing the DAVIS event camera.


\bibliography{arxiv}

\newpage
\appendix
\runninghead{Supplemental Material}{Xue et al. EVENT-BASED NON-RIGID RECONSTRUCTION}

\section{Event Simulation for Non-rigid Object}
\label{sec:event_simulation}

\begin{figure}[H]
\centering

  \begin{subfigure}[b]{0.195\textwidth}
    \includegraphics[width=\textwidth]{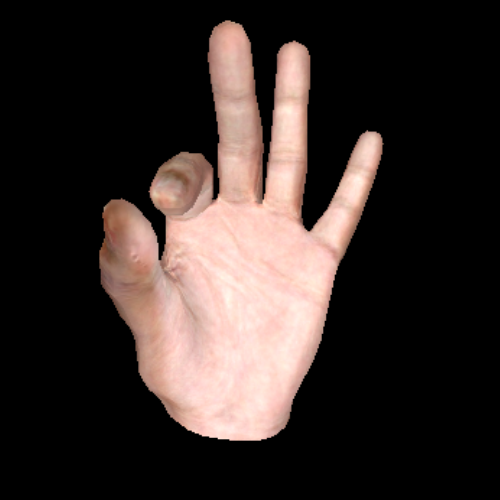}
    \caption{}
    \end{subfigure}
    \hfill
    \begin{subfigure}[b]{0.195\textwidth}
    \includegraphics[width=\textwidth]{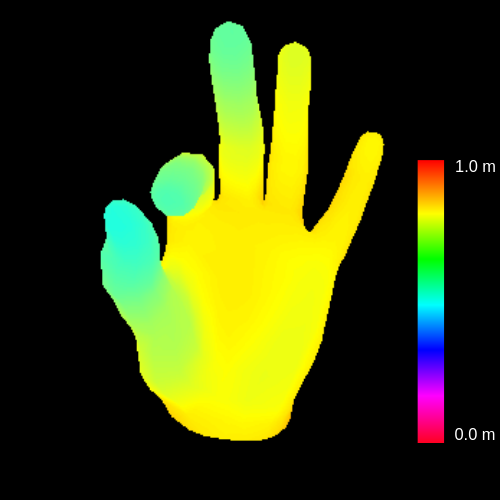}
    \caption{}
    \end{subfigure}
    \hfill
    \begin{subfigure}[b]{0.195\textwidth}
    \includegraphics[width=\textwidth]{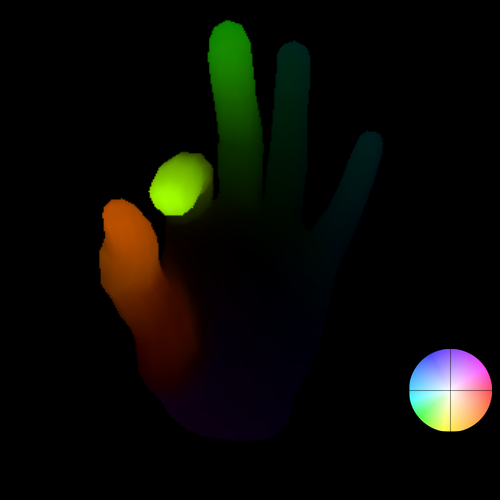}
    \caption{}
    \end{subfigure}
    \hfill
    \begin{subfigure}[b]{0.195\textwidth}
    \includegraphics[width=\textwidth]{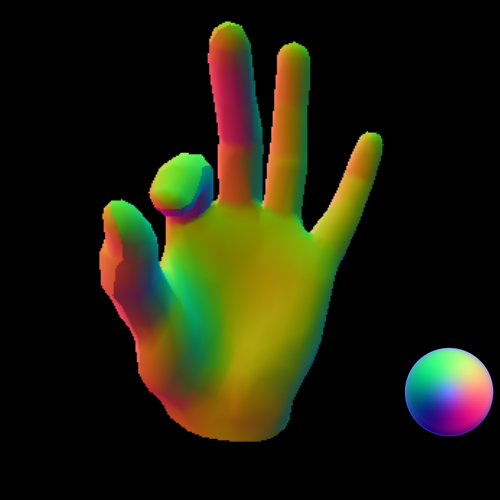}
    \caption{}
    \end{subfigure}
    \hfill
    \begin{subfigure}[b]{0.195\textwidth}
    \includegraphics[width=\textwidth]{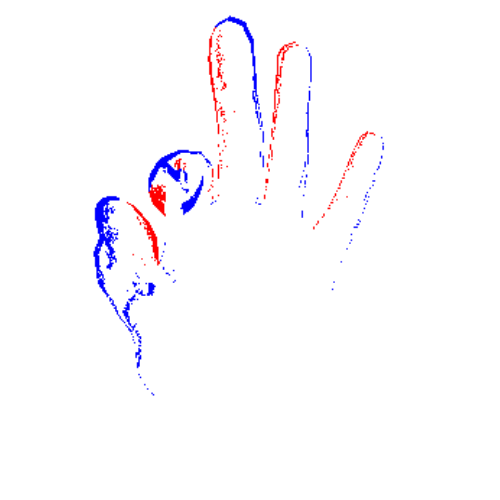}
    \caption{}
    \end{subfigure}
  \caption{All data modalities in our simulator, including (a) RGB image, (b) depth map, (c) motion field, (d) normal map, (e) accumulated events in 1/30 seconds.}
  \label{fig:simulator_modalities}
\end{figure}

We propose an event data simulator which generates synthetic events and other data modalities (Fig.~\ref{fig:simulator_modalities}) of human body motion, especially of hand deformation. In addition to the events stream simulation, our simulator is able to simulate RGB image, depth image, 2D motion field, and normal map. Our event stream simulator is inspired by Nehvi's simulator \cite{nehvi2021_diffevsim}, ESIM \cite{rebeg2018_esim}, and Rudnev's simulator \cite{rudnev2021_eventhands}. It combines advantages of above mentioned simulators. We compare our simulator with these existing event stream simulators at the end of the section.

\subsection{Event Generation Model}
Unlike RGB cameras which capture absolute brightness for each pixels at a fixed frame rate, event cameras record logarithmic pixel-level brightness change asynchronously. To simulate the event at time $t_i$, we calculate the absolute logarithmic brightness at each pixel $\mathbf{u_i}$, denoted as $\mathcal{L}\left(\mathbf{u_{i}}, t_{i}\right) $, and compare with the logarithmic brightness value of the last sampled image at time $t_{i-1}$. The polarity $p$ of the event is

\begin{equation}
    p(\mathbf{u}_i, t_i) = 
    \begin{cases}
      $+1$ & \text{if $\mathcal{L}\left(\mathbf{u}_i, t_{i}\right) - \mathcal{L}\left(\mathbf{u}_i, t_{i-1}\right) \geq C^{+} $ }, \\
      $-1$ & \text{if $\mathcal{L}\left(\mathbf{u}_i, t_{i-1}\right) - \mathcal{L}\left(\mathbf{u}_i, t_{i}\right) \geq C^{-} $ }, \\
    \end{cases}       
\end{equation}

where $C^+$ and $C^-$ are positive and negative contrast threshold, respectively. If the logarithmic brightness change is less than the corresponding contrast threshold, no event is generated at pixel $\mathbf{u}_i$.

To simulate the motion field, we project the 3D movement of each mesh face onto the 2D image plane. Then, we adjust the time interval of next sample according to the largest motion vector magnitude among all pixels. For more details about the adaptive sampling principle, please refer to~ESIM \cite{rebeg2018_esim}. 

We also generate noisy events to make the simulated data more realistic. As in ESIM \cite{rebeg2018_esim}, we sample the contrast threshold from a normal distribution with standard deviation $\sigma$ for each pixel at every sampling step to add uncertainty to the event generation. To simulate salt-and-pepper noise on the background, we sample the probability of each pixel from a uniform distribution in $[0, 1]$, and compare with a predefined threshold. If the probability exceeds the threshold, a noise event is generated. We then sample the timestamp of the noise events uniformly in $[t_{i-1}, t_i]$. For the adjustment of threshold to have the similar amount of salt-and-pepper noise as real event cameras, please refer to Rudnev's simulator \cite{rudnev2021_eventhands}. 

\subsection{Non-Rigid Parametric Models}
When simulating hands alone using MANO \cite{mano_hand_model}, we use the full 45-dimensional PCA parameters of MANO. 
SMPL-X \cite{smplx} is an expressive parametric human model, which models shape and pose of the human body using SMPL \cite{loper2015_smpl}, hand pose using MANO \cite{mano_hand_model}, and facial expression using FLAME \cite{tianye1027_flame, pixie2021Yao}. Note that a PCA low-rank approximation of the pose parameters of the MANO model is used. The body pose is represented by 3-DoF orientations of 21 Joints while facial expression is controlled by 10 PCA parameters in expression space.  We show the simulated data stream of these models in the supplementary file. 

Our simulator takes a sequence of pose parameters of body and hand, the facial expression parameters as well as the simulation time as inputs and simulates event stream, RGB image, depth map, motion field and normal map (see Fig.~\ref{fig:simulator_modalities}). 
Similar as ESIM \cite{rebeg2018_esim}, our simulator assumes that the pose and expression parameters change linearly between two consecutive inputs of the sequence.

\subsection{Comparison}
We compare our proposed simulator with existing event stream simulators ~\cite{nehvi2021_diffevsim, rebeg2018_esim,  rudnev2021_eventhands} in Table \ref{tab:comparison_simulator}. Compared to Nehvi's simulator \cite{nehvi2021_diffevsim}, the simulator we propose can generate the 2D motion field for deforming objects and is accelerated using CUDA. Our experiments show that our simulator is 78-times faster by simulating the same hand motion of MANO model \cite{mano_hand_model}. Compared to ESIM \cite{rebeg2018_esim}, our simulator can simulate events of human body motion. Compared to Rudnev's simulator \cite{rudnev2021_eventhands}, our simulator uses the adaptive sampling strategy to avoid redundancy for small motion, while Rudnev's method samples image frames every 0.001 seconds regardless of the motion.  

\begin{table}[H]
    \centering
    \begin{tabular}{|c|c|c|c|}
    \hline    &  Objects  & MF \& AS  & CUDA\\
    \hline  Rebecq et al. \cite{rebeg2018_esim}  &  Rigid Objects & \cmark    &  \xmark\\
    \hline  Nehvi et al. \cite{nehvi2021_diffevsim} &  MANO \cite{mano_hand_model}  & \xmark   & \xmark \\
    \hline  Rudnev et al. \cite{rudnev2021_eventhands}  &  SMPL-H \cite{mano_hand_model} & \xmark  & \cmark \\
    \hline  Ours &  SMPL-X \cite{smplx}  & \cmark    & \cmark \\
    \hline
    \end{tabular}
    \vspace{3mm}
    \caption{Comparison between our event simulator and other event simulators. MF stands for motion field while AS stands for adaptive sampling.}
    \label{tab:comparison_simulator}
\end{table}

\paragraph{MANO hand}

\begin{figure}[ht]
\centering

  \begin{subfigure}[b]{0.195\textwidth}
    \includegraphics[width=\textwidth]{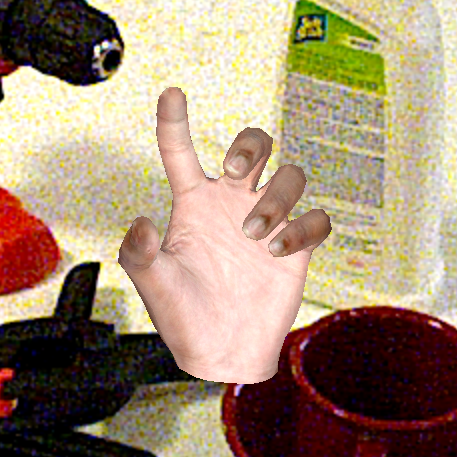}
    \caption{}
    \end{subfigure}
    \hfill
    \begin{subfigure}[b]{0.195\textwidth}
    \includegraphics[width=\textwidth]{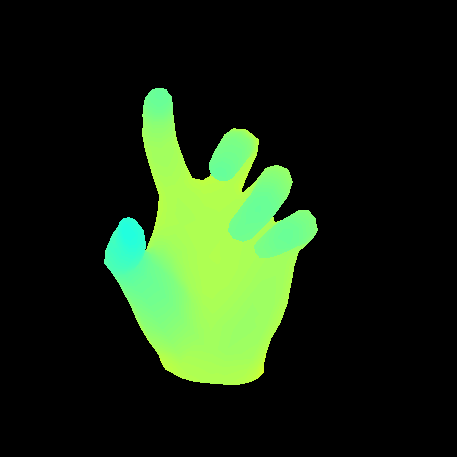}
    \caption{}
    \end{subfigure}
    \hfill
    \begin{subfigure}[b]{0.195\textwidth}
    \includegraphics[width=\textwidth]{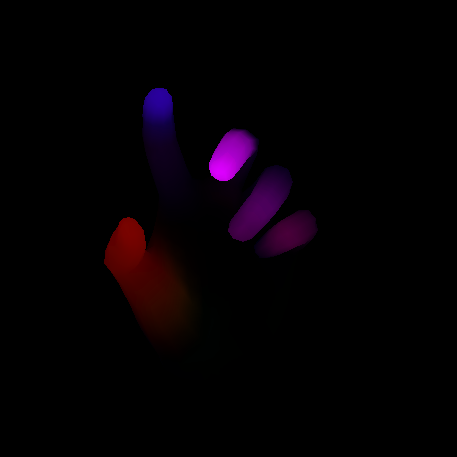}
    \caption{}
    \end{subfigure}
    \hfill
    \begin{subfigure}[b]{0.195\textwidth}
    \includegraphics[width=\textwidth]{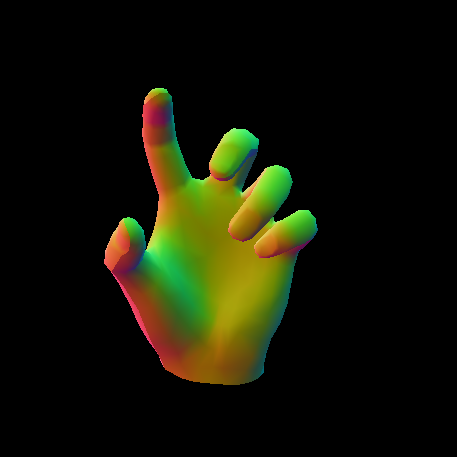}
    \caption{}
    \end{subfigure}
    \hfill
    \begin{subfigure}[b]{0.195\textwidth}
    \includegraphics[width=\textwidth]{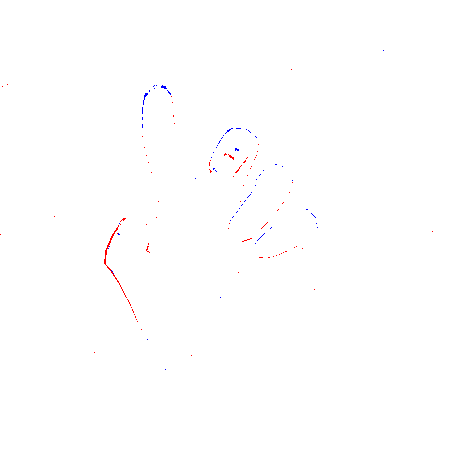}
    \caption{}
    \end{subfigure}
  \caption{All data modalities of MANO~\cite{mano_hand_model} hand model, including (a) RGB image, (b) depth map, (c) motion field, (d) normal map, (e) accumulated events in 1/30 seconds.}
  \label{fig:simulator_modalities_mano}
\end{figure}

The simulated data stream of single MANO~\cite{mano_hand_model} model is shown in Fig.~\ref{fig:simulator_modalities_mano}. Note that the MANO hand model only contains the hand but no arm.

\paragraph{SMPL-X hand}

\begin{figure}[ht]
\centering

  \begin{subfigure}[b]{0.195\textwidth}
    \includegraphics[width=\textwidth]{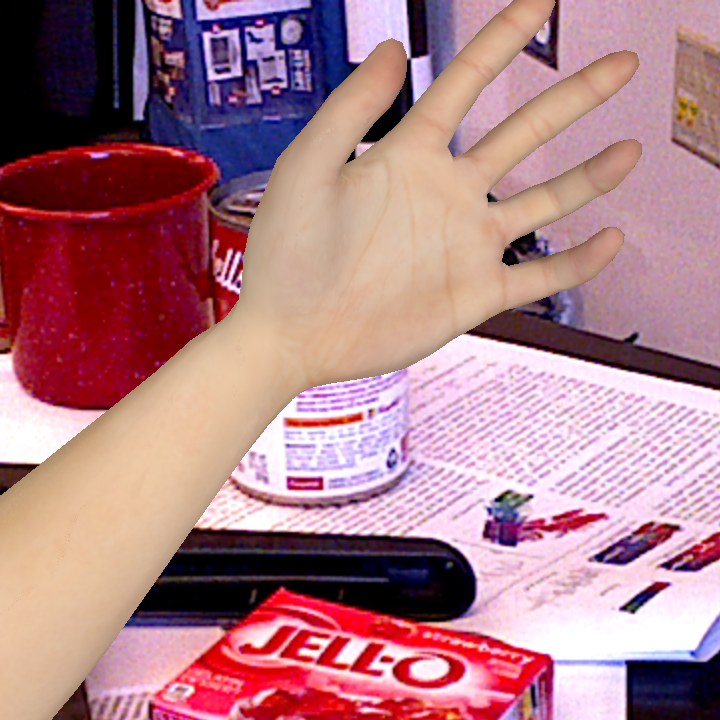}
    \caption{}
    \end{subfigure}
    \hfill
    \begin{subfigure}[b]{0.195\textwidth}
    \includegraphics[width=\textwidth]{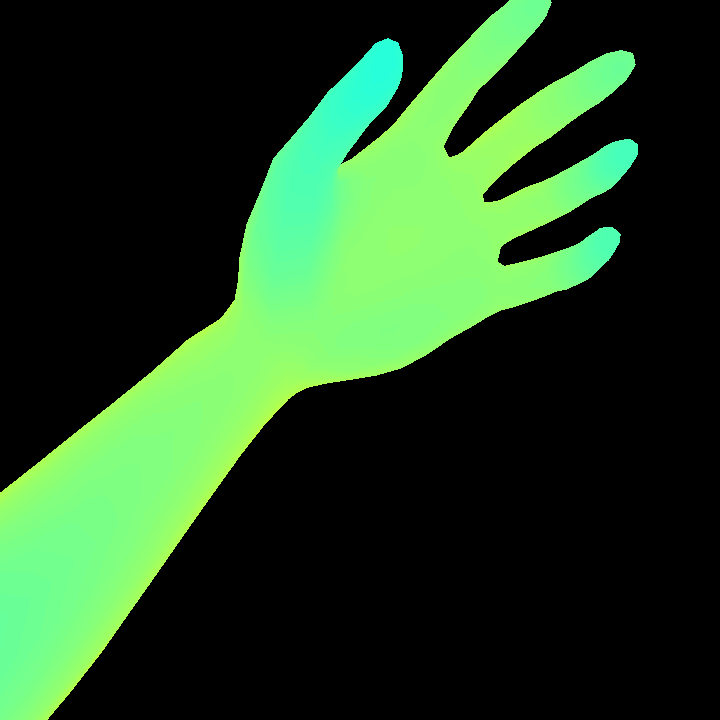}
    \caption{}
    \end{subfigure}
    \hfill
    \begin{subfigure}[b]{0.195\textwidth}
    \includegraphics[width=\textwidth]{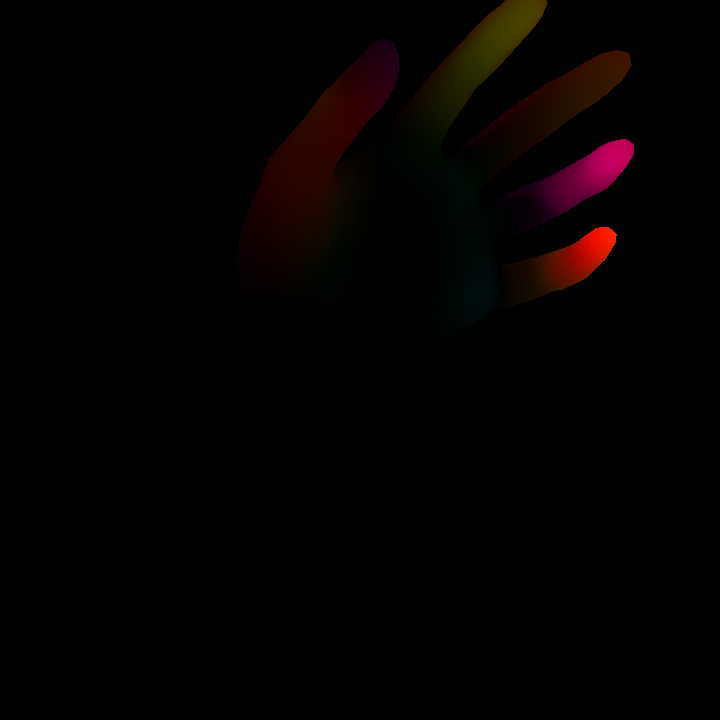}
    \caption{}
    \end{subfigure}
    \hfill
    \begin{subfigure}[b]{0.195\textwidth}
    \includegraphics[width=\textwidth]{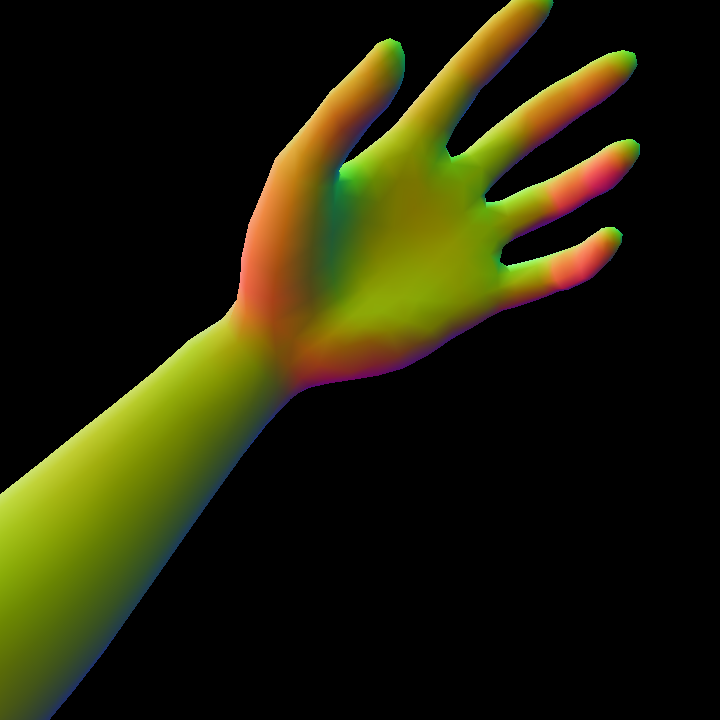}
    \caption{}
    \end{subfigure}
    \hfill
    \begin{subfigure}[b]{0.195\textwidth}
    \includegraphics[width=\textwidth]{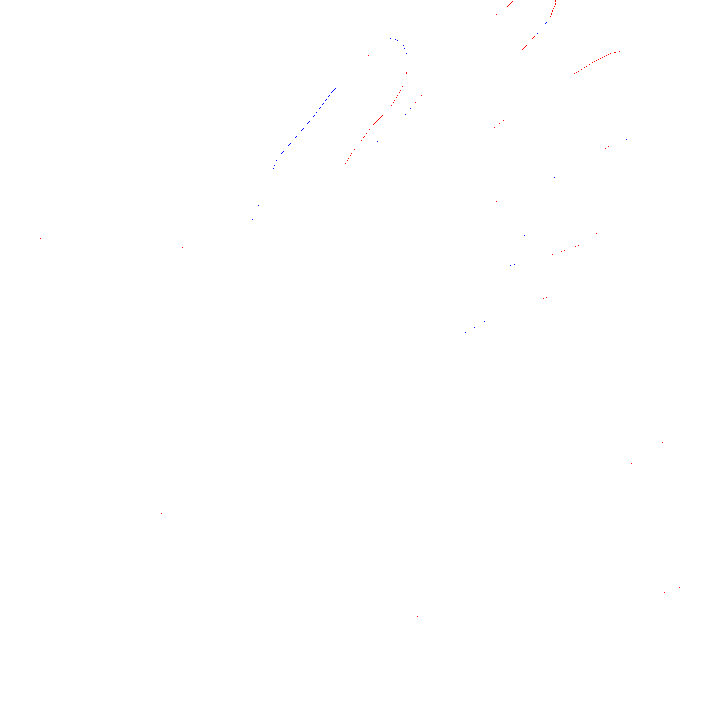}
    \caption{}
    \end{subfigure}
  \caption{All data modalities of SMPL-X~\cite{smplx} hand, including (a) RGB image, (b) depth map, (c) motion field, (d) normal map, (e) accumulated events in 1/30 seconds.}
  \label{fig:simulator_modalities_smplx_hand}
\end{figure}

We visualize the simulated data stream of SMPL-X~\cite{smplx} hand model in Fig.~\ref{fig:simulator_modalities_smplx_hand}. It only contains the motion of the hand. However, attaching the arm to the hand makes it more realistic.

\paragraph{SMPL-X arm and hand}

\begin{figure}[ht]
\centering

  \begin{subfigure}[b]{0.195\textwidth}
    \includegraphics[width=\textwidth]{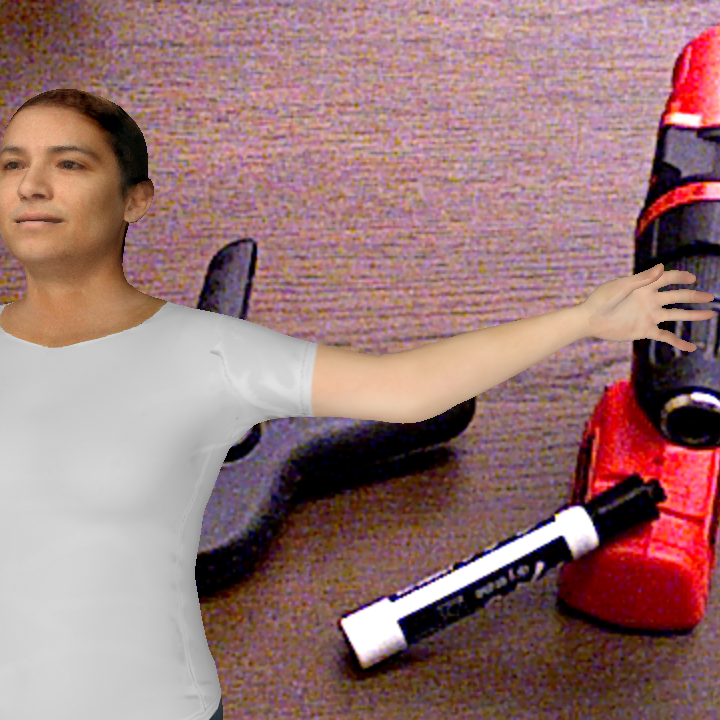}
    \caption{}
    \end{subfigure}
    \hfill
    \begin{subfigure}[b]{0.195\textwidth}
    \includegraphics[width=\textwidth]{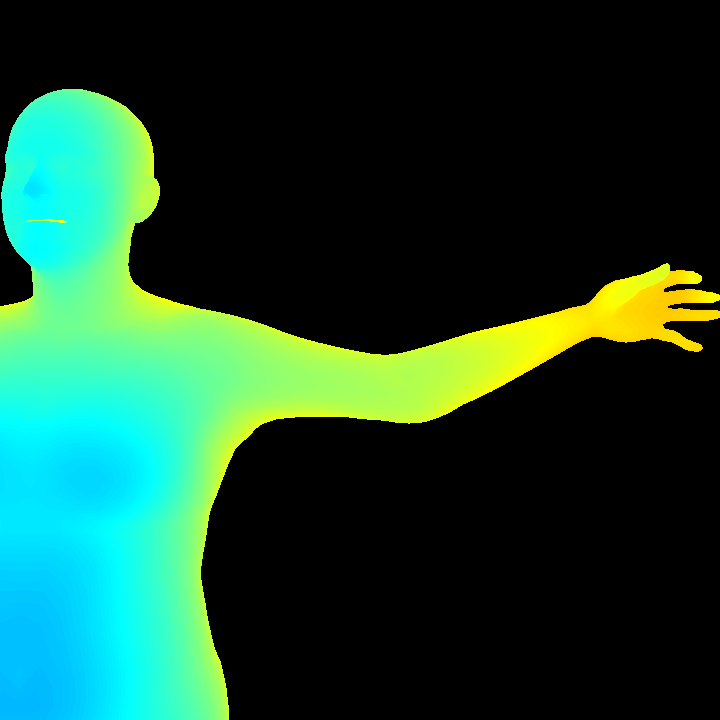}
    \caption{}
    \end{subfigure}
    \hfill
    \begin{subfigure}[b]{0.195\textwidth}
    \includegraphics[width=\textwidth]{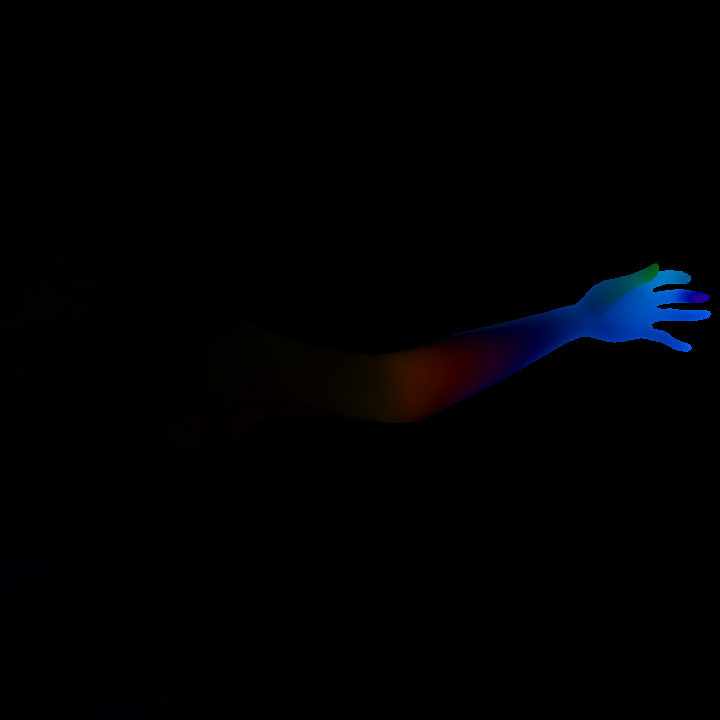}
    \caption{}
    \end{subfigure}
    \hfill
    \begin{subfigure}[b]{0.195\textwidth}
    \includegraphics[width=\textwidth]{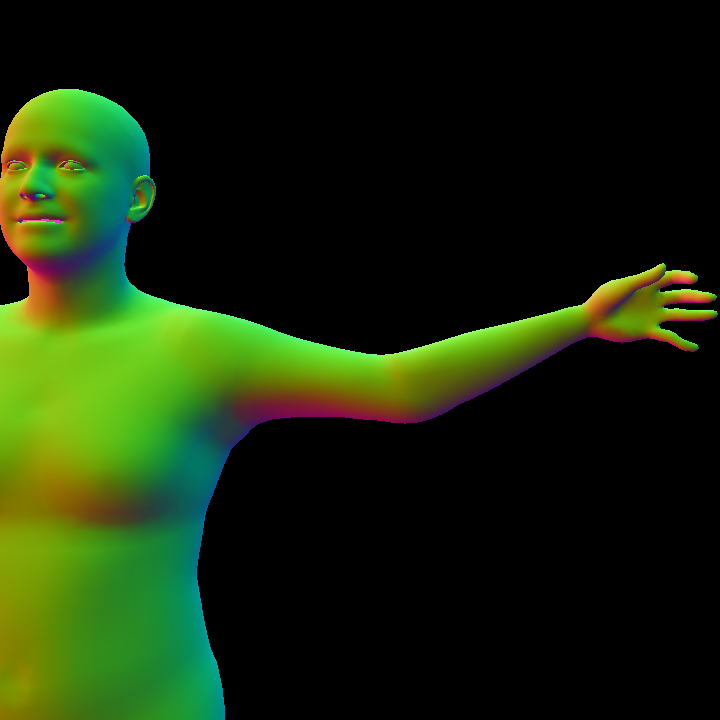}
    \caption{}
    \end{subfigure}
    \hfill
    \begin{subfigure}[b]{0.195\textwidth}
    \includegraphics[width=\textwidth]{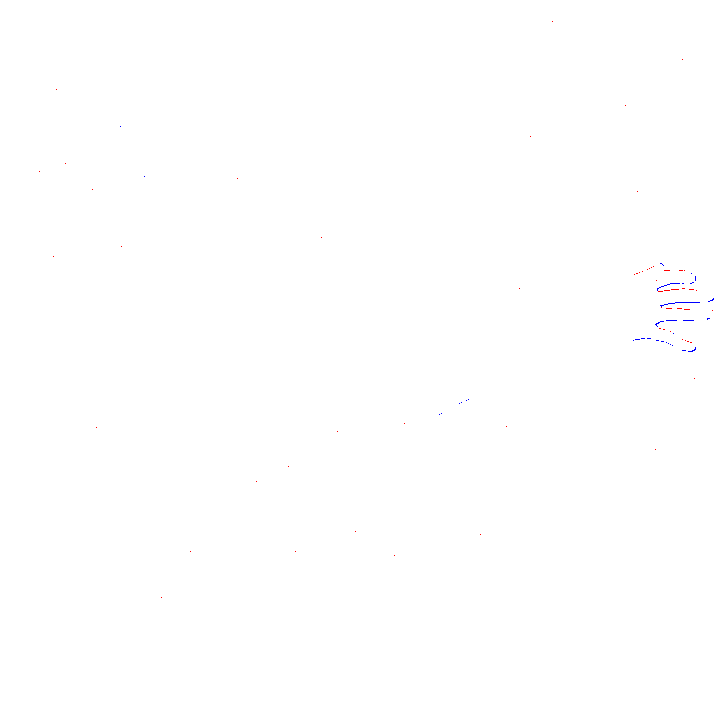}
    \caption{}
    \end{subfigure}
  \caption{All data modalities in SMPL-X~\cite{smplx} arm \& hand, including (a) RGB image, (b) depth map, (c) motion field, (d) normal map, (e) accumulated events in 1/30 seconds.}
  \label{fig:simulator_modalities_smplx_body_hand}
\end{figure}

The simulated data modalities of SMPL-X~\cite{smplx} arm and hand motion are visualized in Fig.~\ref{fig:simulator_modalities_smplx_body_hand}. The data contains the combined motion of the arm and the hand. 

\section{Incremental EM reconstruction using Contour Events}
\subsection{Algorithm}

In this section, we show the pseudo code of our approach in algorithm \ref{algo:em_tracking}. It consists of an E-step to estimate the event association likelihood, and an M-step to maximize the association likelihood.

\begin{algorithm}
\caption{Incremental EM reconstruction using event-based cameras}\label{algo:em_tracking}
\hspace*{\algorithmicindent} \textbf{Input:} events $\{e_k, \cdots, e_{k+N-1}\}$ in spatio-temporal buffer $W_k$\\
\hspace*{\algorithmicindent} \textbf{Output:} optimized mesh pose parameter $\theta_k$  
\begin{algorithmic}[1]
\Procedure{ExpectationMaximization}{}
\State $\theta_k \gets \textit{initialization of mesh pose} $, 
\State \emph{\textbf{E-step}}:
\State $f(\theta_k) \gets \textit{generate mesh model given pose parameter } \theta_k$, 
\State $obj\_{func} \gets 0, \textit{initialization of objective function} $
\For{$e_i$ in $\{e_k, \cdots, e_{k+N-1}\}$}
\State $d^i_{normal} \in [F] \gets \textit{dot product between event } e_i \textit{ to F faces}$, 
\State $d^i_{lateral} \in [F] \gets \textit{lateral distance from event } e_i \textit{ to F faces}$, 
\State $d^i_{longitudinal} \in [F] \gets \textit{longitudinal distance from event } e_i \textit{ to F faces}$, 
\State $P(e_i|a, f(\theta)) \in [F] \gets \textit{Likelihood of event } e_i \textit{ caused by F faces}$,
\State $E(LL(f(\theta_k|e_i, a))) \gets \textit{expectaion of log-likelihood of event } e_i$, 
\State $obj\_func \gets obj\_func + E(LL(f(\theta_k|e_i, a)))$,
\EndFor
\State \emph{\textbf{M-step}}:
\State $\theta_k \gets \underset{\theta_k}{\operatorname{argmax}} \textit{ } obj\_func$
\If {$\textit{Optimization not converged}$}
\State \textbf{goto} \emph{E-step}.
\EndIf
\EndProcedure
\end{algorithmic}
\end{algorithm}


\section{Experiments}

\subsection{Real Data Mesh Template Initialization}
As a template-based method, we assumes the initial pose and shape parameters are known. For real data, we used MeshGraphormer~\cite{meshGraphormer} to infer MANO mesh model from a grayscale image. Then, we minimized the chamfer distance between the predicted hand mesh and the PCA-parametrized MANO hand mesh to optimize shape and pose parameters of the captured hand. Finally, we fix the shape and pose of the hand, manually fine-tune the global rotation and translation of the mesh model by the visual alignment between the rendered 2D hand image and captured hand image.

\subsection{Hyperparameter Tuning}

We use Optuna~\cite{optuna} to tune hyperparameters in our approach and Nehvi's method~\cite{nehvi2021_diffevsim}. 
The hyperparameters in our work comprise sharpness control parameters ($\alpha$, $\beta$, $\gamma$), early stopping threshold in the optimization, expectation update threshold, and the outlier distance threshold.
The hyperparameters in Nehvi's method is the contrast threshold $C$, the smoothness control weight $w$, and weights of individual loss terms. 

For each scenario, we have 10 random training sequences to tune the hyperparameters. 
We use the MPJPE as the metric of the loss function. 
Optuna will find the smallest MPJPE error for the hyperparameters. 
Depending on the scenarios, we have different settings of hyperparameters for the motion reconstruction based on the MANO model and the SMPL-X model.

\subsection{Drift}

\begin{figure}[!htbp]
\centering
\begin{subfigure}{0.45\textwidth}
    \includegraphics[width=\textwidth]{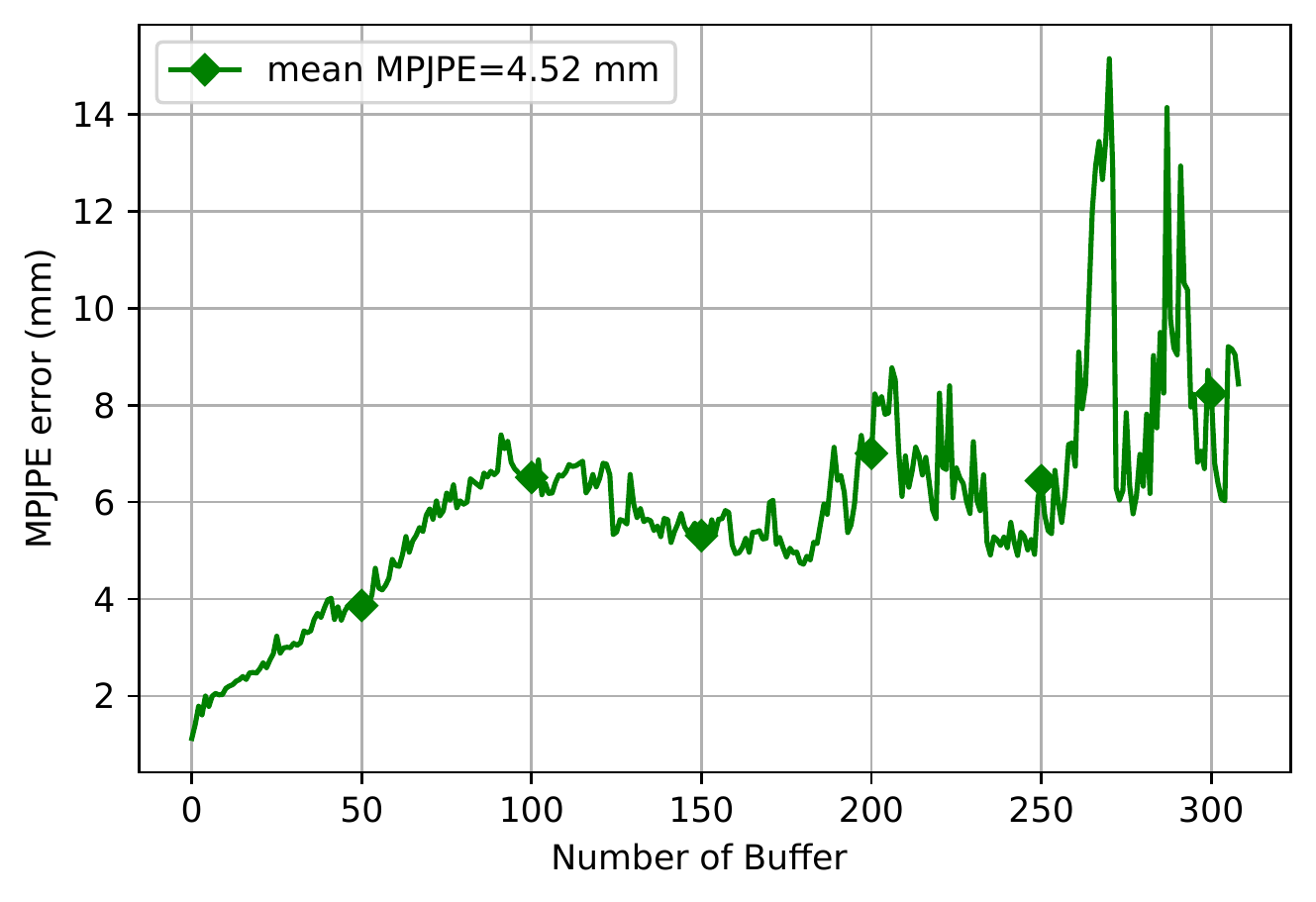}
    \caption{}
\end{subfigure}
\begin{subfigure}{0.45\textwidth}
    \includegraphics[width=\textwidth]{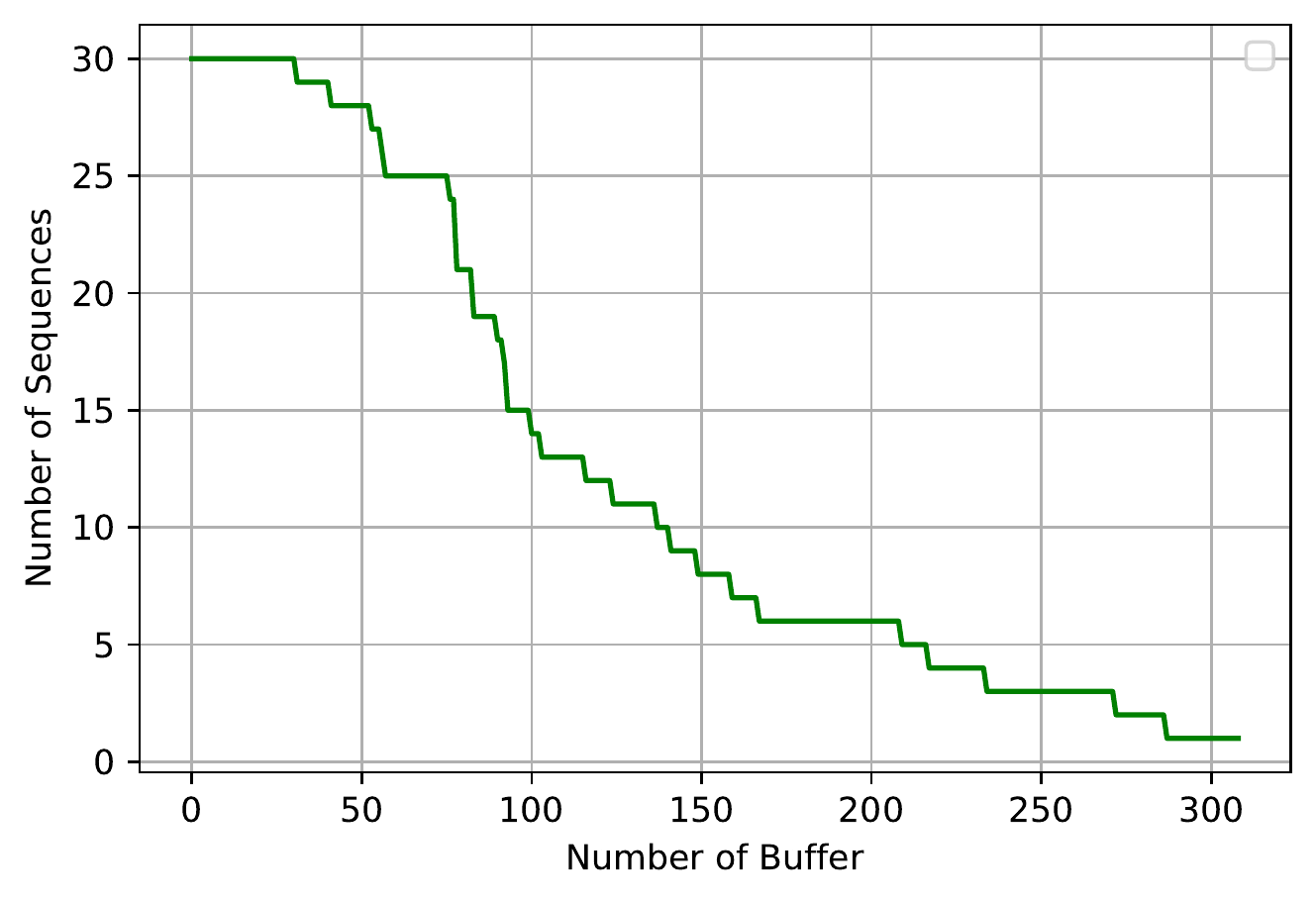}
    \caption{}
\end{subfigure}
\vspace{3mm}
\caption{Drift during optimization of MANO hand reconstruction experiments. (a) Average MPJPE development with the number of processed buffer. (b) Number of sequences still available at the number of processed buffers, indicating over how many sequences the MPJPE in (a) is averaged.}
\label{fig:drift}
\end{figure}

As an incremental optimization-based approach, our approach can also drift, but it can snap the mesh silhouette to the observed events on the contour if sufficient observations are available. Figure \ref{fig:drift} shows that for the MANO hand dataset our approach drifts from the ground-truth initial value at the beginning phase (buffers $(0-100)$), but is able to keep the same level of error in the remaining optimization process.
The sequences in our results are between 0.5s and 2s, while the number of buffers to optimize mainly depends on the speed of the motion.

\subsection{Failure Cases}

\begin{figure}[!htbp]
\centering
\begin{subfigure}{0.32\textwidth}
    \includegraphics[width=\textwidth]{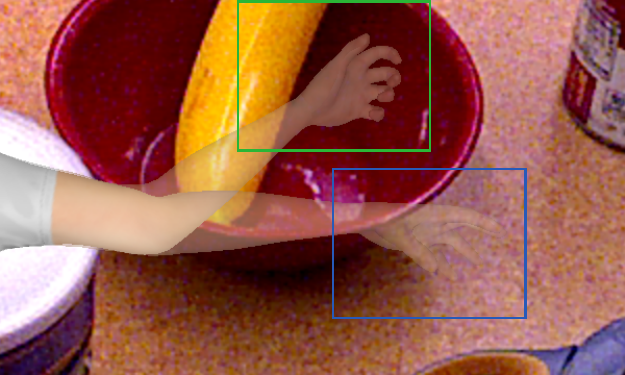}
    \caption{Ground-truth motion in sequence 1.}
    \label{fig:smplx_body_hand_1051_gt}
\end{subfigure}
\hfill
\begin{subfigure}{0.32\textwidth}
    \includegraphics[width=\textwidth]{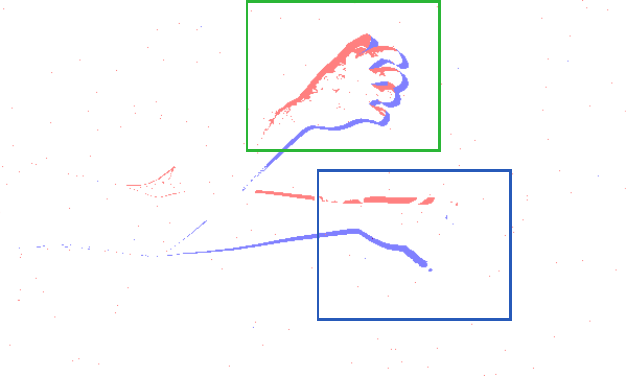}
    \caption{Events in sequence 1.}
    \label{fig:smplx_body_hand_1051_events}
\end{subfigure}
\hfill
\begin{subfigure}{0.32\textwidth}
    \includegraphics[width=\textwidth]{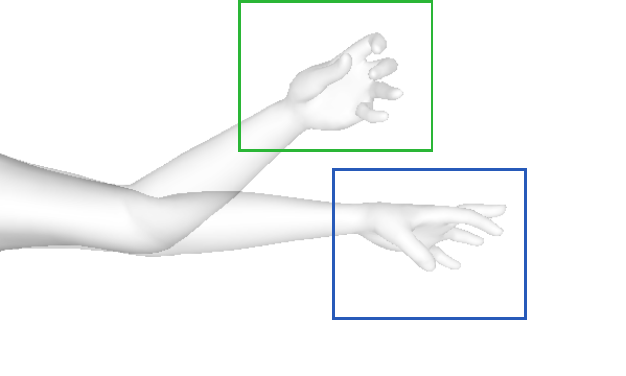}
    \caption{Reconstructed arm and hand pose in sequence 1.}
    \label{fig:smplx_body_hand_1051_rec}
\end{subfigure}
\vspace{3mm}
\caption{Analysis of the failure case of our approach for SMPL-X arm and hand sequences}
\label{fig:smplx_body_hand_1051_analysis}
\end{figure}

Our approach fails in some sequences of SMPL-X body and hand motion. We investigated into why our approach fails in these cases. We visualize the groud-truth images, input event stream, and reconstructed arm and hand in figure \ref{fig:smplx_body_hand_1051_analysis}. The initial pose is in the blue bounding box, and the final pose is in the green bounding box. Figure \ref{fig:smplx_body_hand_1051_events} shows that the hand at the initial pose does not generate valid events. The reason can be inferred from figure \ref{fig:smplx_body_hand_1051_gt}: the fingers at the initial pose has the similar color as the background. According to the event generation model, no events are generated by the motion of fingers. The lack of events leads to the failure case of our approach in this sequence. However, as illustrated in figure \ref{fig:smplx_body_hand_1051_rec}, our approach can still reconstruct the arm motion, because the events of arm motion are generated as usual.

The failure cases due to similar background and object color are more pronounced for the SMPL-X arm \& hand than for SMPL-X hand sequences, because the hand appears smaller in the image and can overlap with the region in the background with similar color more strongly than on the SMPL-X hand sequences.
For example, see Fig.~\ref{fig:smplx_body_hand_1051_analysis}b, where a large part of the events on the hand are missing. 
In the SMPL-X hand sequences, the hand appears larger (for example Fig.~\ref{fig:simulator_modalities_smplx_body_hand}b) and the events are more widely distributed in the image, such that often only parts of the hand are affected and the hand pose is better constrained.

\section{Robustness to noise}
\begin{figure}[htbp]
\centering
\includegraphics[width=0.65\linewidth]{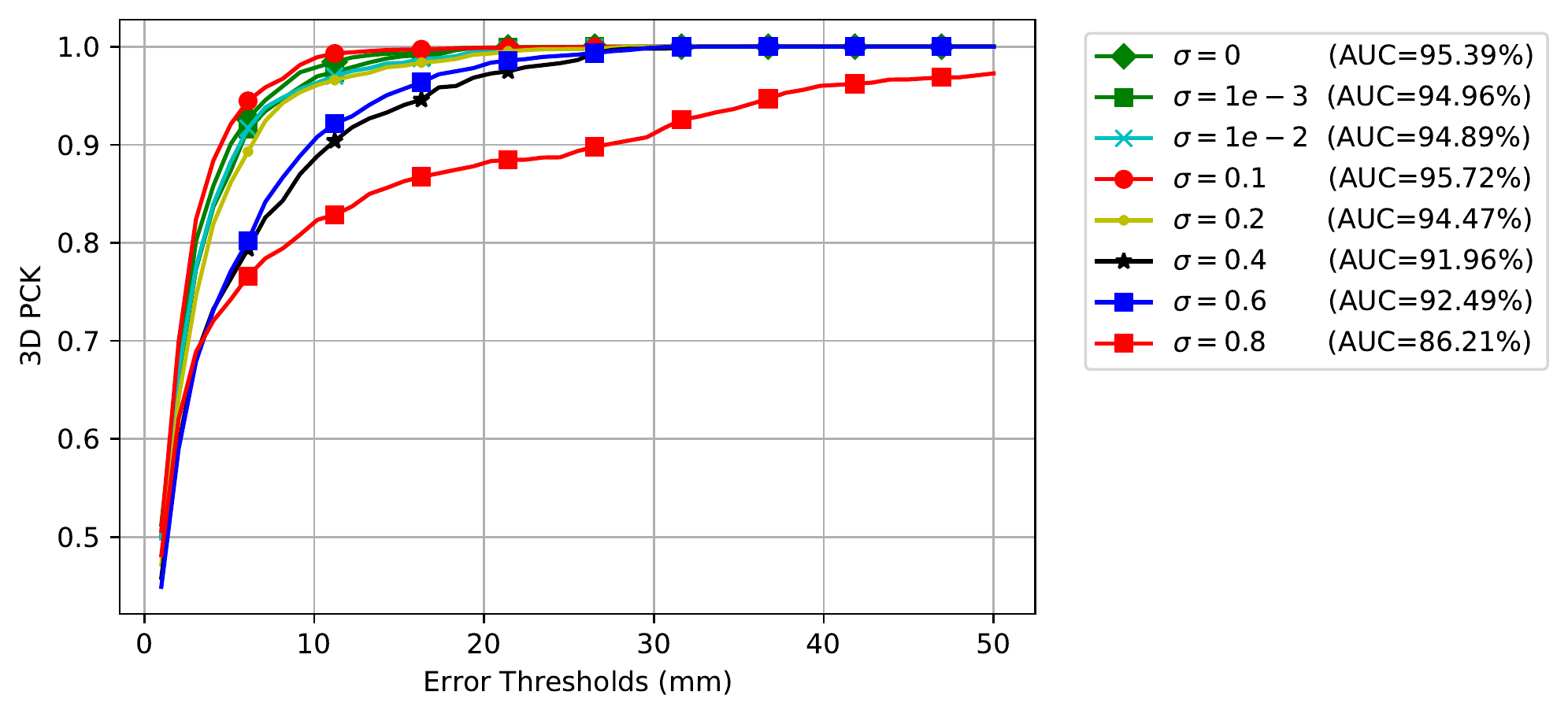}
\caption{Robustness to different level of initial template noise}
\label{fig:robustness_to_init_template}
\end{figure}

\begin{figure}[htp]
    \centering
    \begin{subfigure}{0.2\textwidth}
      \centering
      \includegraphics[width=1.5cm]{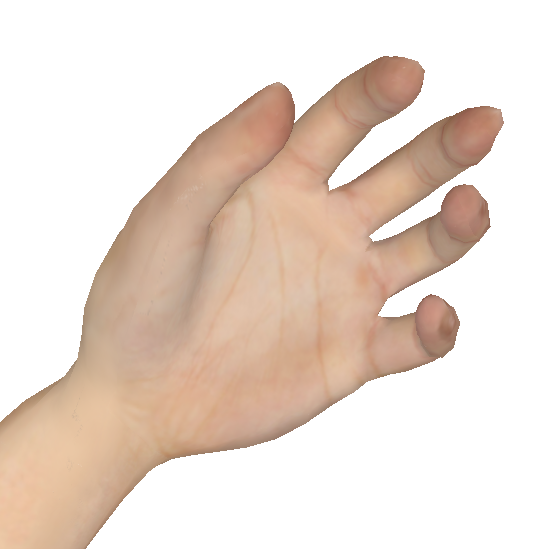}
      \caption{$\theta-0.2 \cdot \mathbf{1}_6$}
    \end{subfigure}%
    \begin{subfigure}{0.2\textwidth}
      \centering
      \includegraphics[width=1.5cm]{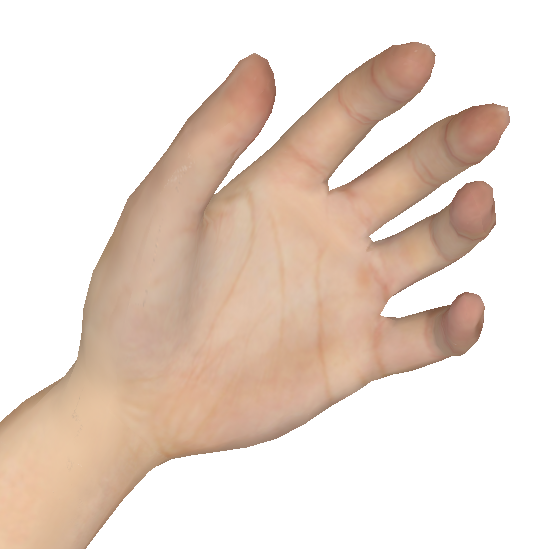}
      \caption{$\theta \in \mathbb{R}^6$}
    \end{subfigure}
    \begin{subfigure}{0.2\textwidth}
      \centering
      \includegraphics[width=1.5cm]{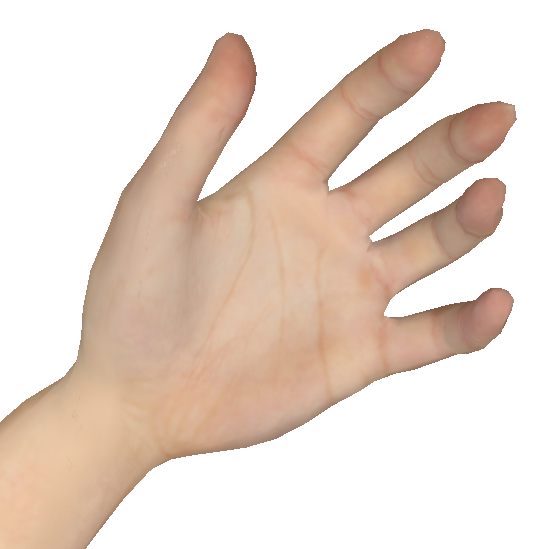}
      \caption{$\theta+0.2 \cdot \mathbf{1}_6$}
    \end{subfigure}
    \vspace*{2ex}
    \caption{Variation of MANO hand parameters}
    \label{fig:mano_hand_variation}
\end{figure}

We evaluate robustness to noisy inputs on the SMPL-X hand motion sequences. In the first experiment, we investigate the robustness to noisy initial templates of objects. Here, we sample 6-dimensional initial pose parameters of hand model from a Gaussian distribution with the mean of ground-truth values and different standard deviations. The 3D-PCK curve and AUC value of each standard deviation are in Fig.~\ref{fig:robustness_to_init_template}. The result demonstrates that our approach still has AUC of $0.86$ when the standard deviation is $0.8$. Note that a noise level of $\sigma = 0.2$ is already high for MANO hand parameters which are in the scale $-2$ to $2$ (see figure~\ref{fig:mano_hand_variation} for hand parameters $\theta \in \mathbb{R}^6$). 

\begin{figure}[ht]
\centering
  \begin{subfigure}[b]{0.45\columnwidth}
  \centering
    \includegraphics[width=\linewidth]{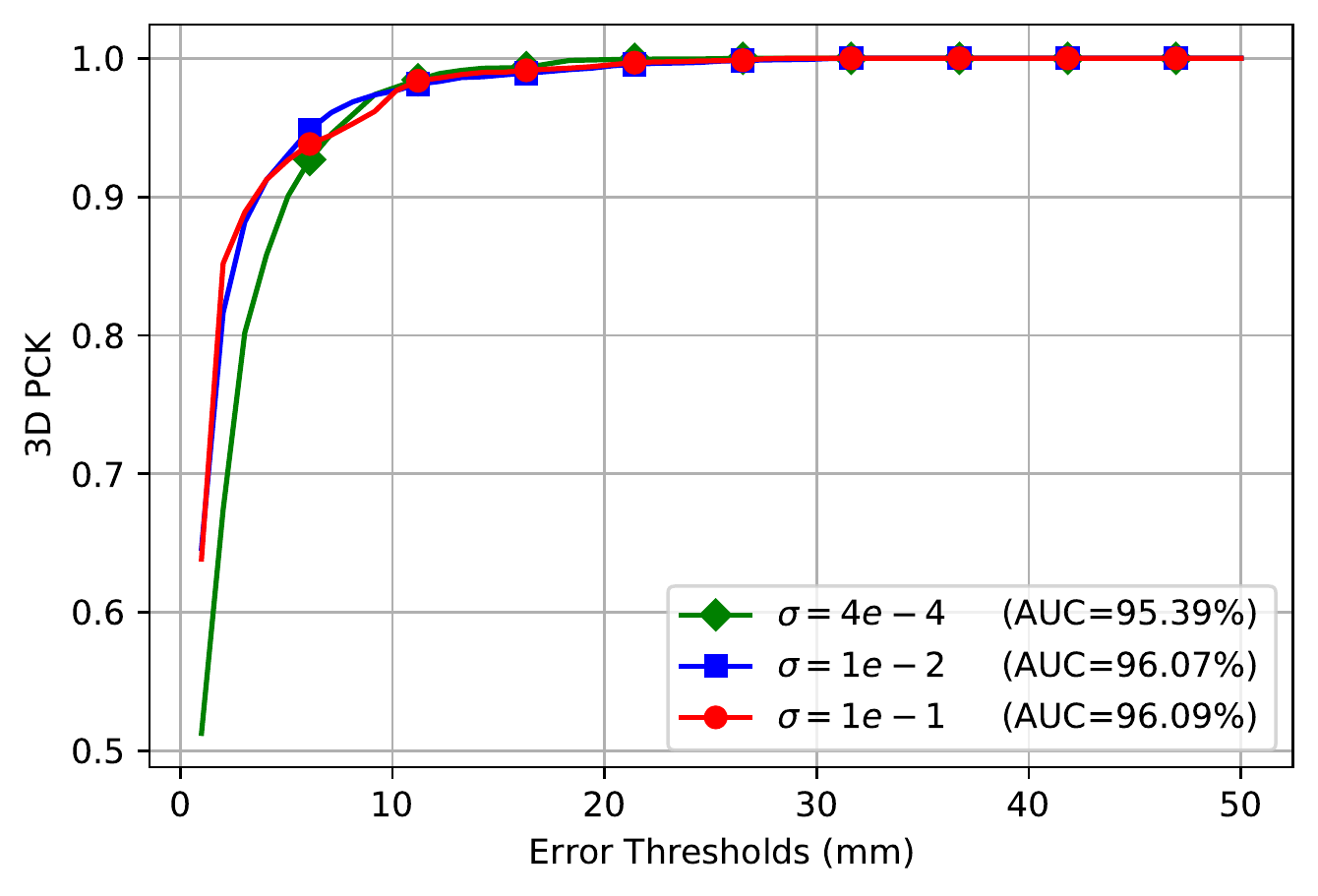}
    \caption{}
    \label{fig:robustness_to_contrast_threshold}
  \end{subfigure}
    \bigskip
  \begin{subfigure}[b]{0.45\columnwidth}
  \centering
    \includegraphics[width=\linewidth]{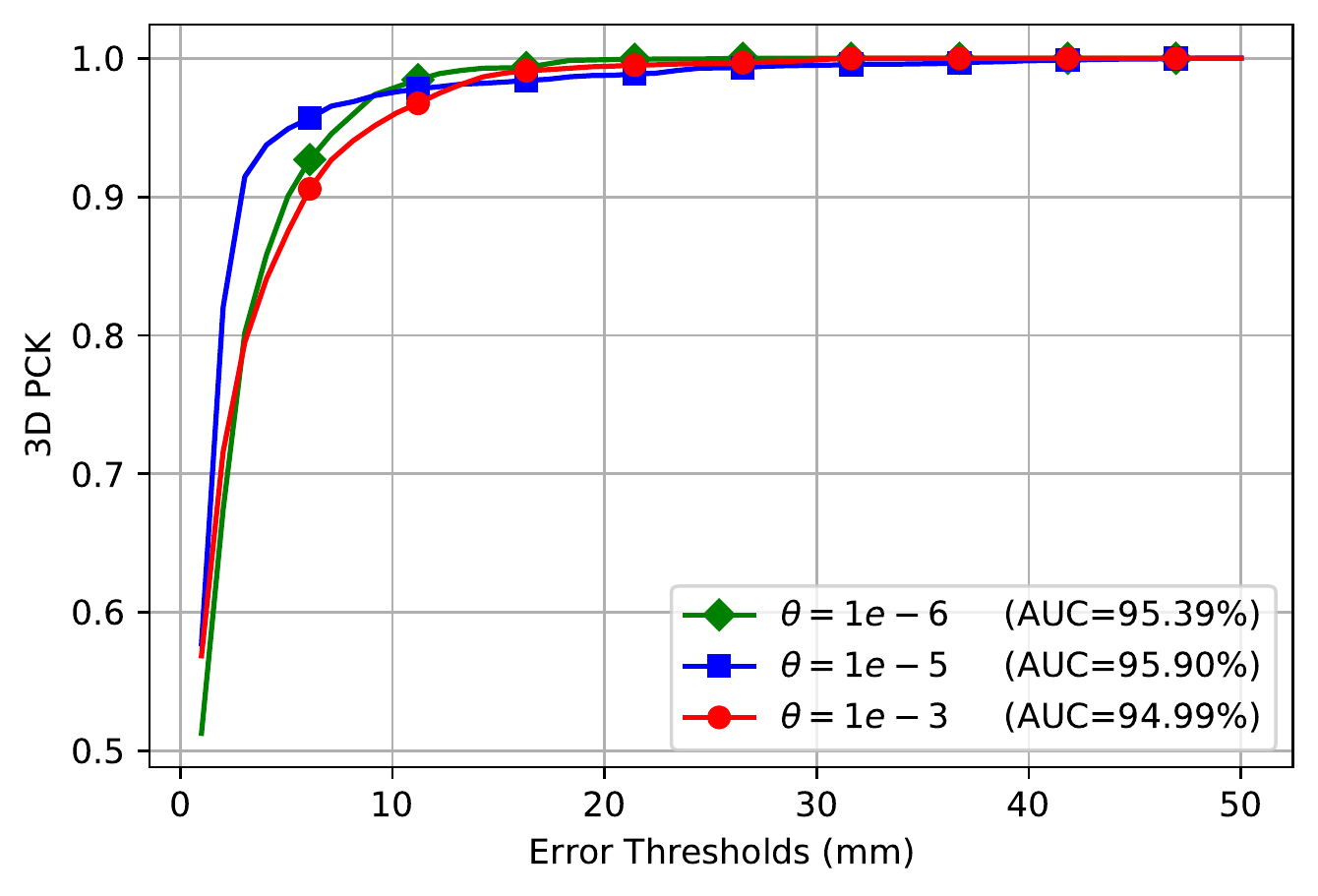}
    \caption{}
    \label{fig:robustness_to_sp_noise}
  \end{subfigure}
  \caption{Robustness to different level of (a) contrast threshold uncertainty; (b) salt-and-pepper noise.}
  \label{fig:robustness_to_noise}
\end{figure}

In the second experiment, we evaluate robustness to noise in the input event stream. 
Noise is caused by the uncertainty of contrast threshold and salt-and-pepper noise in evaluation sequences. Here, we use different levels of standard deviation for contrast threshold sampling and threshold for salt-and-pepper noise to simulate event streams of the different noise levels with the same motion. We show the 3D-PCK curves and AUC values of different noise levels in Fig.~\ref{fig:robustness_to_noise}. 

The result in Fig.~\ref{fig:robustness_to_contrast_threshold} demonstrates that our approach is robust to different levels of uncertainty on contrast threshold in the event generation process. Besides, Fig.~\ref{fig:robustness_to_sp_noise} shows that our approach has solid performance on different amounts of salt-and-pepper noise too. 

\section{Ablation Study}
\subsection{Likelihood Formulation}
In the first ablation study, we investigate variants of the data likelihood term formulated for E-step and M-step on SMPL-X hand motion sequences. The data likelihood of E-step is formulated by the lateral probability, the longitudinal probability, and the contour probability:

\begin{equation}
\begin{split}
P(x_i \mid z_i=j, \boldsymbol{\theta})  \propto P_{lateral} \cdot P_{longitudinal} \cdot P_{contour}.
\end{split}
\label{eq:E3}
\end{equation}

In the ablation study, we formulate the data likelihood in the E-step by either lateral probability and longitudinal probability:
\begin{equation}
    P \left(x_i \mid z_i=j, \boldsymbol{\theta} \right) \propto P_{lateral} \cdot P_{longitudinal},
    \label{eq:E2_longitudinal}
    \end{equation}

or the lateral probability and the contour probability:
\begin{equation}
    P \left(x_i \mid z_i=j, \boldsymbol{\theta} \right) \propto P_{lateral} \cdot P_{contour}.
    \label{eq:E2_normal}
    \end{equation}
    
The proposed data likelihood in the M-step is formulated by the lateral probability and the longitudinal probability:
\begin{equation}
P\left(x_i \mid z_i=j, \boldsymbol{\theta} \right) \propto P_{lateral} \cdot P_{contour}.
\label{eq:M2}
\end{equation}

In the ablation study, we formulate the data likelihood only with the lateral probability:
\begin{equation}
    P \left(x_i \mid z_i=j, \boldsymbol{\theta} \right) \propto P_{lateral}.
     \label{eq:M1}
    \end{equation}
    
 We demonstrate the ablation study in the SMPL-X hand motion reconstruction. The quantitative results of above mentioned variants are shown in Table \ref{tab:ablation_study_data_likelihood}.
 
\begin{table}[h]
\centering
\begin{tabular}{lcc}
\hline 
                                        & MPJPE $(mm)$ & AUC $(\%)$ \\
\hline 
E3M2 (Eq. \ref{eq:E3}, \ref{eq:M2})& $\mathbf{1.5289}$  & $\mathbf{95.9308}$ \\
$\text{E2}_{\text{normal}}$M2 (Eq. \ref{eq:E2_normal}, \ref{eq:M2}) & $1.6523$ & $93.5601$ \\
$\text{E2}_\text{{longitudinal}}$M2 (Eq. \ref{eq:E2_longitudinal}, \ref{eq:M2})& $2.2500$ & $92.7352$ \\
E3M1$_\text{{lateral}}$ (Eq. \ref{eq:E3}, \ref{eq:M1})& $1.9891$ & $92.8573$\\
 \hline
\end{tabular}
\vspace{3mm}
\caption{Ablation Study on probability terms of the data likelihood in the E-step and the M-step}
\label{tab:ablation_study_data_likelihood}
\end{table}

The quantitative results in table above demonstrate that the contour probability is essential for the formulation of the data likelihood term both in the E-step and the M-step. It also indicates that introducing longitudinal probability in the E-step can slightly improve the performance. 
our full data likelihood formulation (Eq. \ref{eq:E3}, \ref{eq:M2}) has best accuracy on the SMPL-X hand motion sequences.

\subsection{Soft and Hard Association}

In the second ablation study, we investigate the soft association and hard association in the M-step on SMPL-X hand motion sequences. For the soft association, we maximize the formulated logarithmic likelihood for all mesh faces in the M-step. For the hard association, we select the mesh face which has the highest probability according to the E-step, and maximize only the likelihood for the mesh face in the M-step.

\begin{table}[h]
\centering
\begin{tabular}{lcc}
\hline 
                                        & MPJPE $(mm)$ & AUC $(\%)$ \\
\hline 
Soft Association  & $1.11$ & $96.38$ \\
Hard Association & $1.19$ & $96.45$ \\
 \hline
\end{tabular}
\vspace{3mm}
\caption{Ablation Study on soft association and hard association}
\label{tab:ablation_study_soft_hard_association}
\end{table}

The result in Table~\ref{tab:ablation_study_soft_hard_association} shows that the soft association is slightly better in MPJPE. Generally, the soft and the hard association does not have huge difference. Our analysis is that the E-step already assigns a relatively high probability to one mesh face. Thus, the soft association and the hard association achieve similar results in this experiment.

\end{document}